\documentclass[essd]{copernicus}

\nolinenumbers

\begin{document}

\title{BenthiCat: An opti-acoustic dataset for advancing benthic classification and habitat mapping}


\Author[1][hayat.rajani@udg.edu]{Hayat}{Rajani} 
\Author[1]{Valerio}{Franchi}
\Author[2]{Borja}{Martinez-Clavel Valles}
\Author[2]{Raimon}{Ramos}
\Author[1]{Rafael}{Garcia}
\Author[1]{Nuno}{Gracias}

\affil[1]{Computer Vision and Robotics Research Institute, University of Girona, Spain}
\affil[2]{Tecnoambiente SLU, Spain}




\runningtitle{BenthiCat - Multimodal Benthic Dataset}
\runningauthor{H. Rajani et al.}

\received{}
\pubdiscuss{} 
\revised{}
\accepted{}
\published{}


\firstpage{1}

\maketitle


\begin{abstract}
Benthic habitat mapping is fundamental for understanding marine ecosystems, guiding conservation efforts, and supporting sustainable resource management. Yet, the scarcity of large, annotated datasets limits the development and benchmarking of machine learning models in this domain. This paper introduces a thorough multi-modal dataset, comprising about a million side-scan sonar (SSS) tiles collected along the coast of Catalonia (Spain), complemented by bathymetric maps and a set of co-registered optical images from targeted surveys using an autonomous underwater vehicle (AUV). Approximately 36000 of the SSS tiles have been manually annotated with segmentation masks to enable supervised fine-tuning of classification models. All the raw sensor data, together with mosaics, are also released to support further exploration and algorithm development. To address challenges in multi-sensor data fusion for AUVs, we spatially associate optical images with corresponding SSS tiles, facilitating self-supervised, cross-modal representation learning. Accompanying open-source preprocessing and annotation tools are provided to enhance accessibility and encourage research. This resource aims to establish a standardized benchmark for underwater habitat mapping, promoting advancements in autonomous seafloor classification and multi-sensor integration.
\end{abstract}


\introduction  
\label{sec:introduction}

Despite the oceans covering more than 70\,\% of our planet's surface, a minuscule fraction of the seabed has been thoroughly explored. Benthic maps provide information about the types of habitats present on the seafloor, including details about the composition of the substrate (e.g., sand, mud, rock) and the diversity and distribution of benthic organisms such as corals, sponges, and other marine life. These maps are not only valuable for understanding marine biodiversity and conservation efforts but also for economic activities. Typically, the surveys for creating such maps are conducted from oceanographic vessels, using different types of acoustic sensors, such as the side-scan sonar (SSS), due to their long range of operation and their ability operate effectively even in poor visibility conditions. This is followed by a labour-intensive process of manual interpretation of the surveyed area by a group of experts, which is not only time-consuming but also economically taxing. Additionally, video samples are often collected over certain surveyed areas to allow a visual confirmation of seafloor characteristics, compensating for the low spatial resolution and low signal-to-noise ratio of acoustic sensing, which would otherwise make it difficult to classify the benthos even by a trained human eye. This, however, further adds to the operational costs of creating such detailed benthic maps.

While AUV-based acoustic mapping has evolved dramatically, there is still a clear need to move the ability to interpret these maps from a post-mission expert-based analysis to an on-mission machine-learning capability. The automation of seafloor classification could revolutionize this domain, offering real-time, precise underwater mapping. In the recent years, there have been various studies on mission-time seafloor classification in SSS imagery \citep{rtseg, dcnet, burguera, r2cnn, convit} using Deep Neural Networks (DNNs). The problem is posed as dense pixel-wise categorization, or semantic segmentation, of different seafloor types in patches of acoustic images. However, with the exception of the study by \cite{burguera}, none of the aforementioned studies have published the data used for their respective experiments. Moreover, due to the high costs and logistical intricacies of data collection in marine environments, compounded by the meticulous and time-consuming nature of data labelling, there is already a severe lack of publicly available labelled datasets that are also large enough for training Machine Learning (ML) models. As such, to the best of our knowledge, there is no standard dataset to date for benchmarking and validation of algorithms or reproduction of results, thereby presenting formidable barriers to stronger statistical analysis and methodological improvements.

To overcome these challenges, this paper introduces a thorough multi-modal dataset for training and benchmarking ML models on seafloor classification in SSS imagery. Our team's collaboration with the Spanish company TecnoAmbiente has yielded an extensive repository of about a million SSS tiles, encompassing a wide variety of seafloor types, collected as part of a large-scale survey along the entire coast of Catalonia. This huge amount of data can be easily leveraged for self-supervised pre-training of ML models. In order to facilitate subsequent fine-tuning of these models, we also manually annotated about 36000 images with segmentation masks. We additionally include the raw and mosaiced SSS waterfalls, together with coarse interpretations, as part of this dataset. Although the emphasis of this work lies on seafloor classification in SSS imagery, we also release bathymetric maps of the surveyed area collected via the onboard Multibeam Echosounder (MBES) together with all the associated navigation information to encourage further research. Furthermore, we conducted an independent series of optical surveys to complement the SSS imagery. These surveys solely targeted specific areas identified during the acoustic surveys to collect samples of certain classes of interest, in order to help alleviate inter-class ambiguities in SSS imagery. This resulted in an additional 178000 images, which were subsequently paired with SSS tiles from the corresponding geographic location, culminating in an opti-acoustic dataset. We believe that such a dataset could be leveraged in modality fusion pipelines to further enhance the quality of benthic classification.

The remainder of this paper is organised as follows. Section~\ref{sec:related_work} presents an overview of other publicly available datasets for benthic mapping. Sections \ref{sec:data_acquisition_a} and \ref{sec:data_acquisition_o} describe the details of the surveys conducted for acoustic and optical data acquisition, including the system setup and sensor characteristics, and outline how the collected data was processed. Section~\ref{sec:dataset}, on the other hand, presents details on the additional processing of the collected data to compile datasets for training ML models. Section~\ref{sec:repository} explains how the datasets are structured. Finally, Section~\ref{sec:conclusion} presents conclusions and directions for future work.


\section{Related Work}
\label{sec:related_work}

Publicly available datasets containing opti-acoustic data, particularly SSS, are notoriously scarce. A notable contribution in this direction is the AURORA multi-sensor dataset for robotic ocean exploration \citep{aurora}, which was compiled from multiple missions using the Autosub 6000 AUV by the National Oceanography Centre (NOC) in the Greater Haig Fras Marine Conservation Zone (MCZ). This dataset is designed to facilitate the testing of control, navigation, and sensor processing algorithms for AUVs, as well as to serve as a benchmark for research in robotic ocean exploration. It integrates various modalities such as SSS, MBES, and optical images, along with AUV navigation data and other associated sensor metadata. However, the dataset does not specifically target seabed classification and lacks fine-grained annotations of seabed sediments.

In contrast, the dataset released in conjunction with the work of \cite{burguera} is explicitly focused on seabed segmentation in SSS images. The data was acquired using an EcoMapper AUV in Port de Sóller, Mallorca, Spain, traversing over 4\,\unit{km}, and resulting in a total of 22438 swaths. These swaths were translated to overlapping patches of size $83 \times 83$, annotated across three classes: rocks, sand ripples and mud for training.

On the other hand, the recently published SeafloorAI dataset \citep{seafloorai} is designed to support AI-based seafloor mapping. The dataset includes around 696000 sonar images and 827000 segmentation masks, collected from 62 surveys across 9 regions, covering a total area of over 17300\,\unit{km^2}. The dataset combines SSS and MBES data and provides pixel-wise labels for five types of geological features: sediment type, physiographic zone, habitat, fault, and fold. In addition, SeafloorAI also includes a vision-language extension called SeafloorGenAI, which provides image-level captions and about seven million question–answer pairs, enabling the development of multimodal models. While SeafloorAI provides annotations for seabed sediments, it only focuses on four broad categories such as rock, gravel, sand and mud, and its objectives quite differ from our work. Where we focus on seabed sediment segmentation with an emphasis on fine-grained, high-resolution distinctions relevant for robot perception, SeafloorAI is designed primarily for large-scale geological interpretation. Its broader goal is to support automated mapping of multiple seafloor attributes (including physiographic zones, structural features like faults and folds, and general habitat types) making it well-suited for regional-scale geoscientific studies, but less so for targeted applications for AUVs such as real-time scene understanding, close-range intervention tasks,  navigation and path planning.

\begin{figure*}[!ht]
    \centering
    \includegraphics[width=0.96\textwidth]{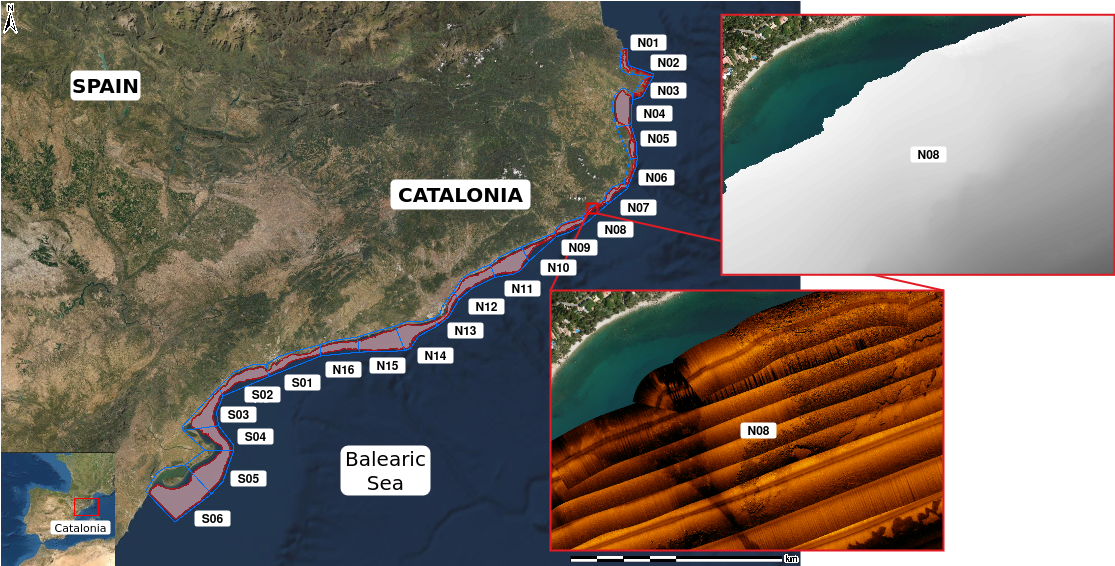}
    \caption{Overview of the surveyed area, divided into 22 sectors, illustrating a portion of the mosaiced SSS transects from sector N08 and the corresponding DEM. The basemap was provided via Esri ArcGIS using OpenStreetMap data. For more information, visit \url{https://www.arcgis.com/home/item.html?id=b834a68d7a484c5fb473d4ba90d35e71} (last access: 15 July 2026).}
    \label{fig:survey_area}
\end{figure*}

To the best of our knowledge these are the only publicly available datasets that align closely with our target application. Other existing datasets in the literature primarily emphasize on object detection \citep{seabed_objects, marine_pulse} or binary segmentation of discrete targets such as shipwrecks \citep{ai4shipwrecks}, pipelines \citep{subpipe}, or mines \citep{mines_dataset}, rather than continuous benthic habitats.


\section{Acoustic Data Acquisition}
\label{sec:data_acquisition_a}

The acoustic data presented in this work was collected by Tecnoambiente SLU through various geophysical surveys of the entire coast of Catalonia\footnote{This data was used to create an interactive marine habitat map by the Government of Catalonia, available at \url{https://sig.gencat.cat/visors/habitats_marins.html}}. These surveys involved over 3500 hours of geophysical (SSS) and hydrographic (MBES) data acquisition, covering more than 1900\,\unit{km^2} in area, extending from the coastline up to six nautical miles offshore. In order to ensure accurate characterization of the habitat, additional surveys were carried out collecting about 500 sediment samples, and 150 scientific dives and ROV-enabled missions for acquisition of video samples. It is important to note that the goal of these additional surveys was to provide a visual validation of the habitat in certain locations, and not to conduct an extensive optical survey. As such, there was not an easy way to co-register the acoustic and optical data, since the latter comprised much smaller and separate transects. Therefore, the data presented as part of these surveys do not include any optical imagery.

Figure~\ref{fig:survey_area} outlines the surveyed area. The entire coast of Catalonia was subdivided into 22 sectors: 16 along the northern and 6 along the southern coast. Each sector comprised multiple overlapping transects, collecting over 50 million swaths, encompassing a wide variety of seabed types, including sand ripples, mud, posidonia (dead and alive), cymodocea, artificial reefs, rocks, corals, detrital funds, etc. Additionally, miscellaneous objects such as buoys, anchors, conduits and cables, shipwrecks, and agricultural infrastructure were also observed. A total of 26 different classes were identified, which were then grouped into 12 categories based on the similarity of visual features or benthic type as summarized in Appendix~\ref{sec:appendixA}.

The remainder of this section provides further details on the system setup, sensor configurations, processing of the raw data and the methodology for creating annotations.

\subsection{System Setup}
\label{sec:system_setup_a}

\begin{figure}[!ht]
    \centering
    \includegraphics[width=0.47\textwidth]{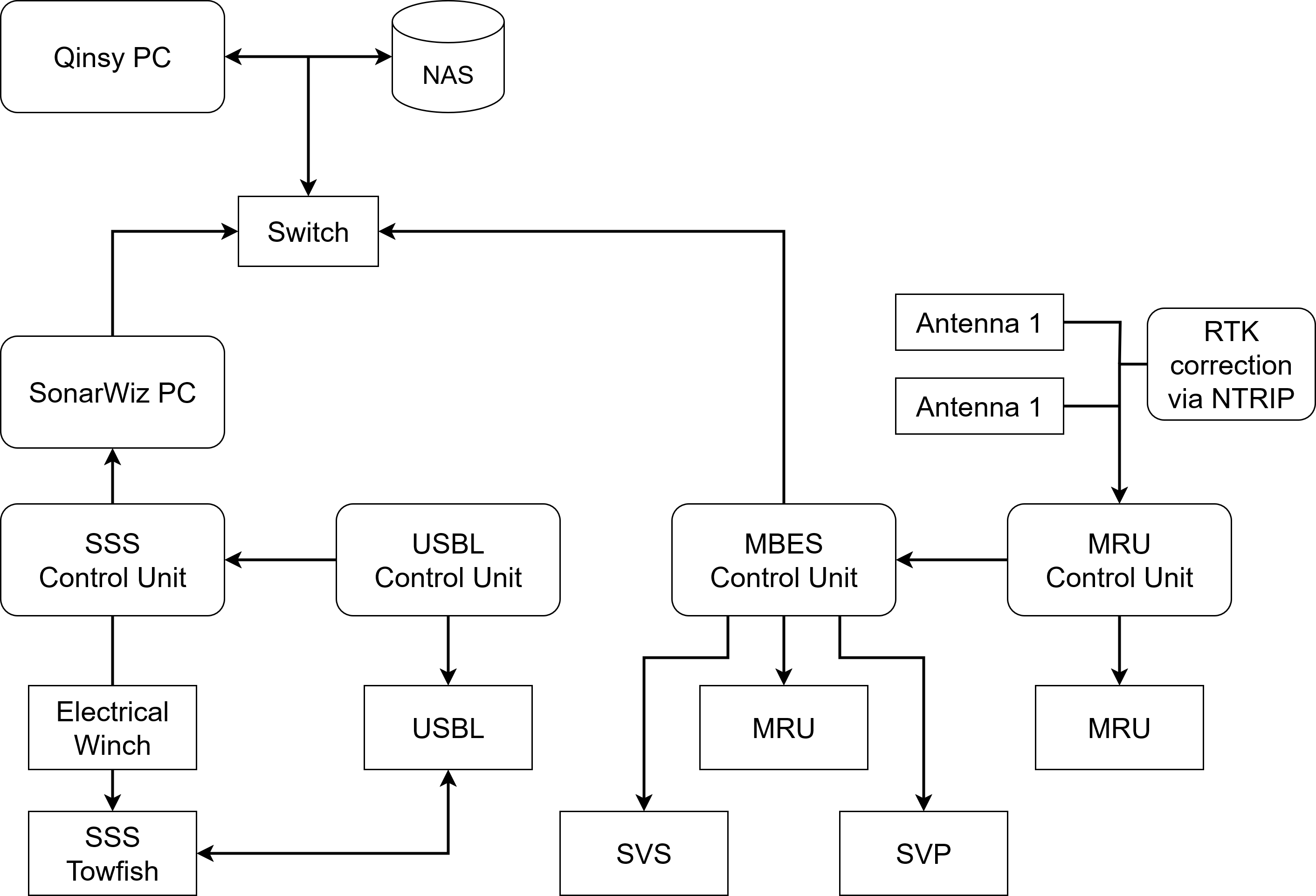}
    \caption{Schema of the equipment on board the survey vessel.}
    \label{fig:sensor_schematic}
\end{figure}

The surveys were conducted using Tecnoambiente's 16\,\unit{m} long survey vessel, \textit{Festina Lente}. The vessel was equipped with a Klein 3000H SSS and an R2Sonic 2022/2024 MBES for acoustic data gathering, together with a Sound Velocity Sensor (SVS) and a Sound Velocity Profile (SVP) for sound velocity corrections, and an Inertial Navigation System (INS), a Motion Reference Unit (MRU) and an Ultrashort Baseline (USBL) to aid with navigation. Figure~\ref{fig:sensor_schematic} depicts a schematic of how these sensors interacted with each other during the course of the surveys.

The Klein 3000H \citep{klein} is a dual frequency (500\,\unit{kHz}\,/\,900\,\unit{kHz}) side-scan sonar that was operated as a towfish hauled behind the survey vessel by a cable controlling its operating depth. Only the scans from the 500\,\unit{kHz} frequency domain were utilized to achieve the best balance between resolution and coverage area per transect. Data acquisition was performed at an altitude varying between 5\,\unit{m} and 15\,\unit{m}, relative to about 10\,\% of the sensor range. The sensor range, on the other hand, was manually adjusted for each zone based on the depth. It varied between 50\,\unit{m} and 150\,\unit{m} for each side (port and starboard. Corrections for navigation errors were applied recursively by the on-board INS during the course of the surveys, together with an online georeferencing of the swaths via the USBL. Initially, a Sonardyne Scout Plus USBL \citep{scoutplus} was used in conjunction with an Navsight Ekinox INS \citep{ekinox}, resulting in a position accuracy of 0.2\,\% of traveled distance and a heading accuracy of 0.02\,\unit{\degree}. This configuration was later replaced by a GAPS Gen 4 USBL \citep{ixblueUSBL}, which combines high performance USBL and a fiber optic INS in the same housing, resulting in a position accuracy of 0.06\,\% of the slant range and a heading accuracy of 0.01\,\unit{\degree} secant latitude.

The R2Sonic 2022 \citep{r2sonic2022} is a wideband shallow water multibeam echosounder that was mounted to the hull of the survey vessel. Data acquisition was performed at an operating frequency of 400\,\unit{kHz}, with a maximum ping rate of 60\,\unit{Hz} and 256 beams per ping. The swath angle was manually adjusted for each zone depending on the depth, never exceeding 140\,\unit{\degree}. The R2Sonic 2022 was later replaced by a R2Sonic 2024 \citep{r2sonic2024}, maintaining the same configurations. For sound velocity corrections, a Valeport miniSVS \citep{miniSVS} was mounted at the head of the MBES. The SVS was operated at an acoustic frequency of 2.5\,\unit{MHz} and a sampling rate of 60\,\unit{Hz}. The 100\,\unit{mm} sensor reported a maximum theoretical error of $\pm$0.017\,\unit{m\,s^{-1}} with a 0.001\,\unit{m} resolution. This was further complemented by sound velocity profiling via a Valeport SWiFT SVP \citep{swiftSVP}, carried out approximately every 3 hours or when changing zones. Real-time corrections for draft and tide variations as well as attitude were performed via the survey vessel's MRU, a Coda Octopus F-175 \citep{codaF175} with RTK corrections, resulting in a heading accuracy of 0.1\,\unit{\degree} and roll and pitch accuracy of 0.03\,\unit{\degree}. This was later replaced by the Navsight Ekinox INS.

Table~\ref{tab:sensor_configs_acoustic_surveys} summarizes the key configurations and characteristics of all sensors.

\begin{table}[!h]
    \small\sf\centering
    \caption{Sensor configurations and characteristics for acoustic surveys.}
    \begin{tabular}{ll}
        \hline
        \textbf{Parameter}&\textbf{Value}\\
        \hline
        \\
        \textbf{Klein 3000H SSS} \\
        Operating Frequency & 500\,\unit{kHz} \\
        Beam Tilt ($\theta$) & 15\,\unit{\degree} \\
        Beam Width ($\alpha$) & 40\,\unit{\degree} \\
        \\
        \textbf{Sonic 2022/2024 MBES} \\
        Operating Frequency & 400\,\unit{kHz} \\
        Resolution & 1\,\unit{cm} \\
        Beam Count & 256 \\
        Beam Angle & 140\,\unit{\degree} (max) \\
        \\
        \textbf{Valeport miniSVS} \\
        Acoustic Frequency & 2.5\,\unit{MHz} \\
        Sample Rate & 60\,\unit{Hz} \\
        Resolution & 0.001\,\unit{m} \\
        Velocity Accuracy & $\pm$0.017\,\unit{m\,s^{-1}} \\
        \\
        \textbf{Valeport SWiFT SVP} \\
        Range & 1375\textendash1900\,\unit{m\,s^{-1}} \\
        Resolution & 0.001\,\unit{m\,s^{-1}} \\
        Accuracy & $\pm$0.02\,\unit{m\,s^{-1}} \\
        \\
        \textbf{Coda Octopus F-175 MRU} \\
        Heading accuracy & 0.1\,\unit{\degree} (with RTK) \\
        Roll and pitch accuracy & 0.03\,\unit{\degree} (with RTK) \\
        \\
        \textbf{Navsight Ekinox INS} \\
        Position accuracy & 0.2\,\% of traveled distance \\
        Heading accuracy & 0.02\,\unit{\degree} \\
        Roll and pitch accuracy & 0.02\,\unit{\degree} \\
        \\
        \textbf{GAPS Gen 4 USBL} \\
        Position accuracy & 0.06\,\% of slant range \\
        Heading accuracy & 0.01\,\unit{\degree} secant latitude \\
        Roll and pitch accuracy & 0.01\,\unit{\degree} \\
        \\
        \hline
    \end{tabular}
    \label{tab:sensor_configs_acoustic_surveys}
\end{table}

\subsection{Data Processing}
\label{sec:data_processing_a}

The raw SSS data were recorded in the eXtended Triton Format (XTF) \citep{xtf2020}, a commonly used open source format for recording hydrographic data, which were subsequently gain-corrected and mosaiced using SonarWiz \citep{sonarWiz}. The mosaiced SSS waterfalls were then georeferenced and interpretations of various sea bottom types were manually drawn by experts using ArcGIS \citep{ArcGIS}. This lead to a coarse benthic habitat map of the entire coast.

\begin{figure}[!h]
    \centering
    \includegraphics[width=0.47\textwidth]{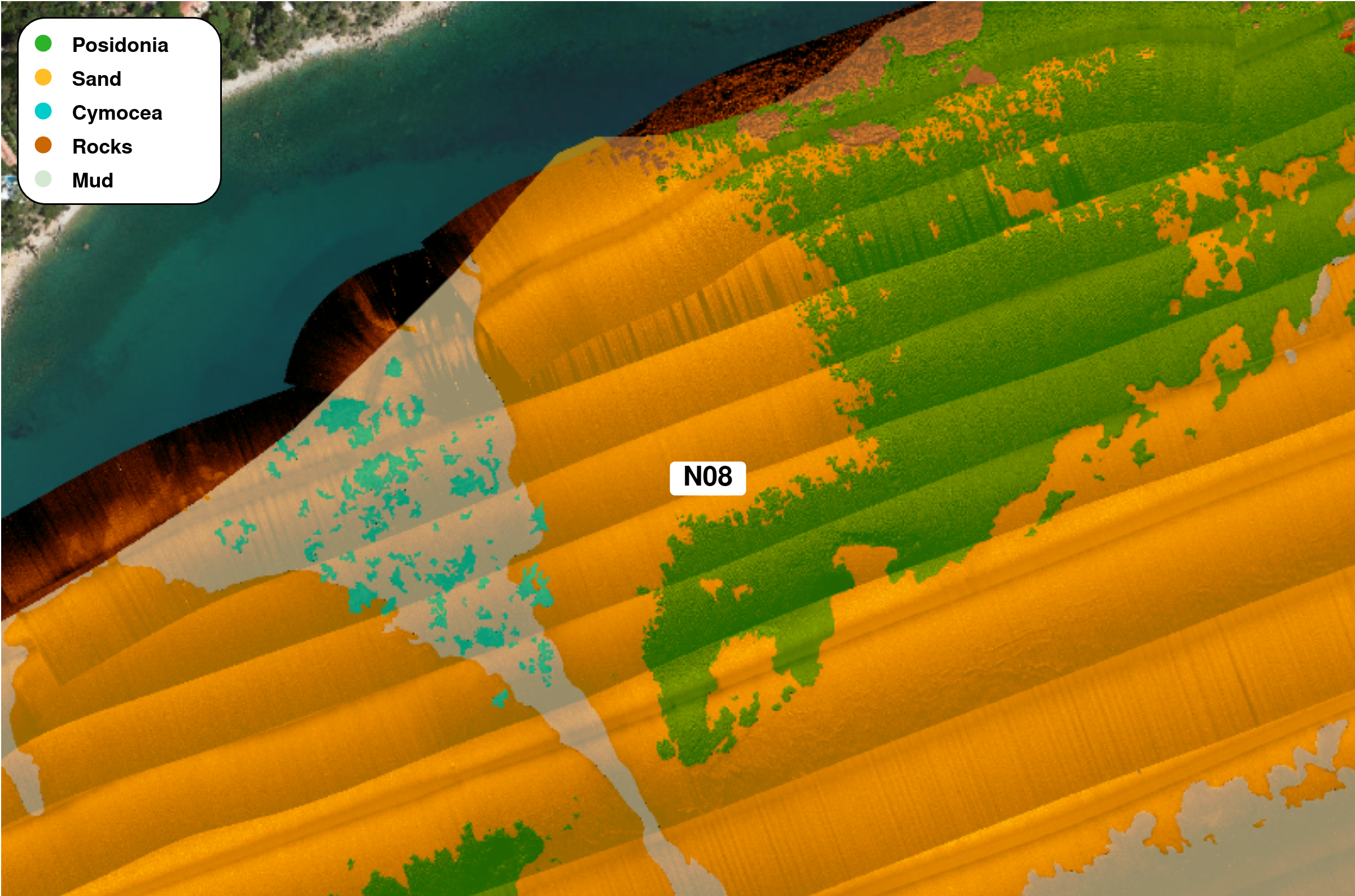}
    \caption{Illustrative example of a portion of an SSS mosaic from sector N08 overlaid with the corresponding benthic habitat interpretation. The basemap was provided via Esri ArcGIS using OpenStreetMap data. For more information, visit \url{https://www.arcgis.com/home/item.html?id=b834a68d7a484c5fb473d4ba90d35e71} (last access: 15 July 2026).}
    \label{fig:annotated_mosaic}
\end{figure}

Figure~\ref{fig:annotated_mosaic} depicts examples of a portion of one such SSS mosaic overlaid with the corresponding interpretation. Although done by experts, the interpretations are prone to human errors. While certain areas lacked the respective interpretations altogether, those of several others were inaccurate as highlighted in the figure. Moreover, further noise was inevitably introduced during the ground truth generation process as detailed in Sect.~\ref{sec:dataset}.

The raw MBES data were recorded using Qinsy \citep{qinsy}, a survey planning and real-time hydrographic data processing software. Qinsy ingests all the data from the acoustic and navigation sensors to compute corrections in real-time and produces a single DB file, a proprietary database format. These raw database files from all the transects were subsequently processed in Qimera \citep{qimera} and converted to the more standard ASCII gridded XYZ files, which were then rasterized using ArcGIS. The so produced Digitized Elevation Models (DEMs) were exported as GeoTIFFs and Shapefiles (.shp) for each sector, and also as a non-editable PDF file containing all cartographic information.


\section{Optical Data Acquisition}
\label{sec:data_acquisition_o}

As previously mentioned, acoustic sensing suffers from low spatial resolution and low signal-to-noise ratio, particularly in dynamic environments, making it difficult to classify the benthos even by a trained human eye. Video samples are, therefore, typically acquired over the surveyed areas to allow a visual confirmation of seafloor characteristics, and thus have a ground-truth for the manual interpretations. Consequently, there is a strong need to enable conducting integrated surveys by fusing information from different sensors, especially using autonomous robotic platforms. This becomes particularly crucial for distinguishing between seabed types with similar appearance in the acoustic images.

As opposed to the acoustic surveys, optical data acquisition was performed at an altitude varying between 3.5\,\unit{m} and 4.5\,\unit{m} depending on the area and weather conditions. Consequently, the varying fields-of-view and the differences in scale between the acoustic and optical images resulted in the lack of one-to-one correspondence between the two modalities. The methodology thus adopted for aligning the data is discussed in Sect.~\ref{sec:dataset}.

\begin{figure*}[!ht]
    \centering
    \includegraphics[width=0.96\textwidth]{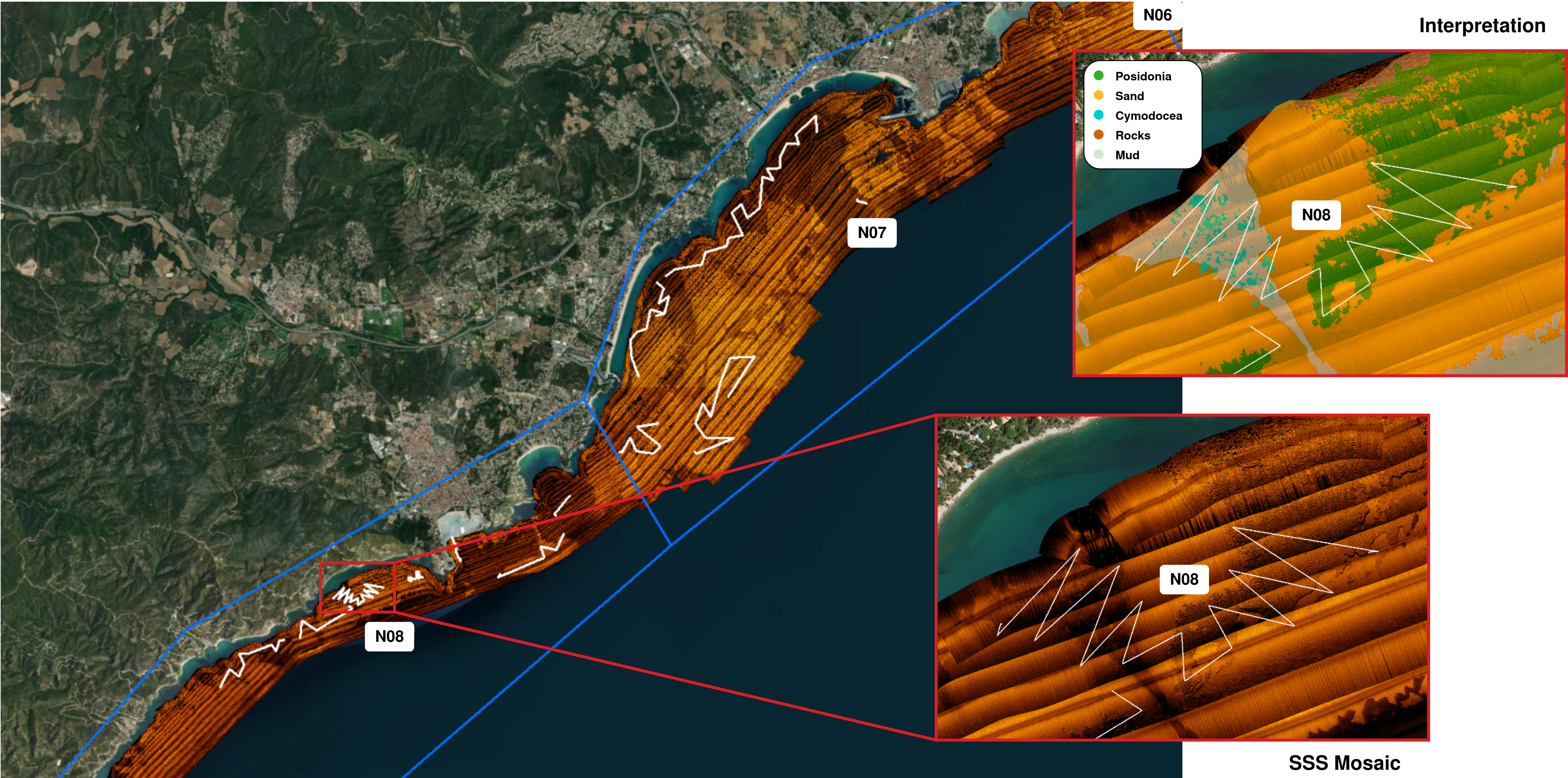}
    \caption{Overview of the optical transects, depicted as white lines, across the surveyed areas in sectors N07 and N08. The basemap was provided via Esri ArcGIS using OpenStreetMap data. For more information, visit \url{https://www.arcgis.com/home/item.html?id=b834a68d7a484c5fb473d4ba90d35e71} (last access: 15 July 2026).}
    \label{fig:optical_transects}
\end{figure*}

As such, we conducted an independent series of optical surveys targeting certain areas from sectors N7 and N8 which were identified during the acoustic surveys to be abundant in the following six classes of interest: fine sediments, sand ripples, rocks, corals, \textit{Posidonia oceanica} and \textit{Cymodocea nodosa}. Figure~\ref{fig:optical_transects} depicts a few transects over the surveyed areas. The surveys were conducted from the 22${}^{\text{nd}}$ until the 26${}^{\text{th}}$ of April 2024, covering approximately a distance of 27.55\,\unit{km} and an area of 118.2\,\unit{km^2} with the optical data acquired at a maximum altitude of 5\,\unit{m}. During the course of the optical surveys, the transects were carried out at three different altitudes: 3.5\,\unit{m}, 4\,\unit{m}, 4.5\,\unit{m} and at two different AUV speeds: 0.5\,\unit{m\,s^{-1}} and 0.8\,\unit{m\,s^{-1}}. Table~\ref{tab:image_footprints} summarises the footprint of each image on the seabed for each of those altitudes for a downward-looking camera.

\begin{table}[!h]
    \small\sf\centering
    \caption{Altitudes and image footprints during the optical surveys.}
    \begin{tabular}{ccc}
        \hline
        \textbf{Altitude} & \textbf{Image Footprint} \\
        \hline
        3.5\,\unit{m} & 10.69\,\unit{m^2} \\
        4.0\,\unit{m} & 13.95\,\unit{m^2}\\
        4.5\,\unit{m} & 17.68\,\unit{m^2}\\
        \hline
    \end{tabular}
    \label{tab:image_footprints}
\end{table}

The remainder of this section provides further details on the system setup, sensor configurations and processing of the raw data.

\subsection{System Setup}
\label{sec:system_setup_o}

\begin{figure}[!hb]
    \centering
    \includegraphics[width=0.4\textwidth]{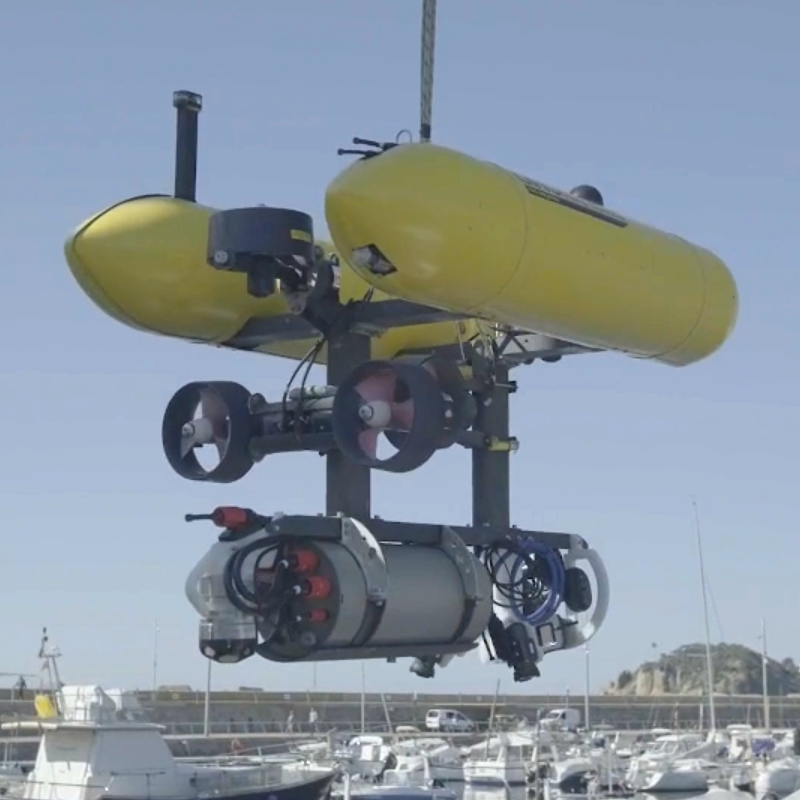}
    \caption{Image of the Girona1000 AUV at port before the start of the optical surveys.}
    \label{fig:girona1000}
\end{figure}

The surveys were conducted using the Girona1000, a variant of the Girona500 hovering AUV \citep{Girona500} developed at the University of Girona for operations at depths of up to 1000\,\unit{m}. The reconfigurable vehicle was equipped with two cameras to gather the optical data, along with a Doppler-Velocity Logger (DVL), a Global Navigation Satellite System (GNSS), an Inertial Navigation System (INS), and a Sound Velocity Sensor (SVS) to estimate the vehicle's position and orientation during the surveys. Figure~\ref{fig:girona1000} displays the Girona1000 AUV with all the mounted sensors, while Tables \ref{tab:girona_sensor_distance_offsets} and \ref{tab:girona_sensor_angle_offsets} list the sensor offsets with respect to the Girona1000's centre of gravity. Table~\ref{tab:sensor_configs_optical_surveys}, in turn, summarizes the key configurations and characteristics of all the sensors.

\begin{table}[!hb]
    \small\sf\centering
    \caption{Sensor distance offsets ($x,y,z$) with respect to the Girona1000's centre of gravity for optical surveys.}
    \begin{tabular}{lcc}
        \hline
        \textbf{Sensor} & \textbf{Offset} (x,y,z) [\unit{cm}]  \\
        \hline
        Phins Compact C3 & $(24.90,37.70,-28.40)$ \\
        DVL1000 & $(-65.0,0.0,49.96)$ \\
        L86 & $(-43.522,-35.005,-65.681)$  \\
        miniSVS & $(-57.05,35.0,-42.725)$  \\
        Blackfly S (left) & $(56.0,0.0,47.0)$  \\
        Blackfly S (right) & $(56.0,0.0,47.0)$ \\ 
        \hline
    \end{tabular}
    \label{tab:girona_sensor_distance_offsets}
\end{table}

\begin{table}[!hb]
    \small\sf\centering
    \caption{Sensor orientation offsets (roll $\psi$, pitch $\theta$, yaw $\phi$) with respect to the Girona1000's centre of gravity for optical surveys. The cameras' orientations were changed multiple times during the surveys based on the area.}
    \begin{tabular}{lcc}
        \hline
        \textbf{Sensor} & \textbf{Offset} ($\psi$,$\theta$,$\phi$) [\unit{rad}] \\
        \hline
        Phins Compact C3 & $(0,3.141593,0)$ \\
        DVL1000 & $(0,0,3.141593)$ \\
        L86 & $(0,0,0)$ \\
        miniSVS & $(0,0,0)$ \\
        \hline
    \end{tabular}
    \label{tab:girona_sensor_angle_offsets}
\end{table}

\begin{table}[!h]
    \small\sf\centering
    \caption{Sensor configurations and characteristics for optical surveys.}
    \begin{tabular}{ll}
        \hline
        \textbf{Parameter} & \textbf{Value} \\
        \hline
        \\
        \textbf{Phins Compact C3 INS} \\
        Position accuracy & 0.3\,\% of traveled distance \\
        Heading accuracy & 0.15\,\unit{\degree} \\
        Roll and pitch accuracy & 0.05\,\unit{\degree} \\
        \\
        \textbf{DVL1000} \\
        Maximum altitude & 75\,\unit{m} \\
        Accuracy & $\pm$0.1\,\%$\pm$0.1\,\unit{cm\,s^{-1}} \\
        Velocity resolution & 0.01\,\unit{m\,s^{-1}} \\
        Maximum ping rate & 8\,\unit{Hz} \\
        \\
        \textbf{miniSVS} \\
        Acoustic Frequency & 2.5\,\unit{kHz} \\
        Sample Rate & 1\textendash32\,\unit{kHz} \\
        Resolution & 0.001\,\unit{m} \\
        Velocity Accuracy & $\pm$0.20\,\unit{m\,s^{-1}} \\
        \\
        \textbf{Teledyne FLIR Blackfly S GigE} \\ 
        Megapixels & 2.8\,\unit{MP} \\
        Maximum Frame Rate & 43\,\unit{Hz} \\
        Gain Range & 0\textendash47\,\unit{dB} \\ 
        \\
        \hline
    \end{tabular}
    \label{tab:sensor_configs_optical_surveys}
\end{table}

The Phins Compact C3 \citep{ixbluePhinsCompact}, based on fiber-optic gyroscope (FOG) technology, is a compact INS designed for customised sub-sea integration. It was used onboard the vehicle as the core Inertial Measurement Unit (IMU) to acquire linear accelerations and angular velocities from the raw sensor measurements. The DVL1000-4000m \citep{DVL10004000}, rated for operational depths of up to 4000\,\unit{m}, was used to measure the vehicle's linear velocity, whereas the Valeport miniSVS \citep{miniSVS} supplied low noise, high accuracy and high resolution sound velocity and pressure readings using digital Time-of-Flight processing. On the other hand, a Quectel L86 GNSS module \citep{L86}, equipped with a patch antenna, was mounted on the antenna mast to provide surface positioning. All sensors were interfaced with the Phins Compact C3, which fused their measurements through its commercial INS algorithm consisting of an Unscented Kalman Filter (UKF) \citep{ukf}, providing the real-time pose and velocity estimates of the vehicle throughout the survey.

Two FLIR Blackfly S GigE cameras \citep{FLIR} were mounted on the bottom hull of the Girona1000 AUV. The cameras were specifically arranged to maximise the overlap between the optical data and the near-field region of the sonar data, with each camera oriented toward opposite sides of the vehicle (see Fig.~\ref{fig:girona1000}, bottom-center). The cameras feature a Sony IMX429 sensor model with a 2$/$3\,\unit{inch} format, a pixel size of 4.5\,$\mu$\unit{m}, and a native resolution of $1936 \times 1464$ (2.8\,\unit{MP}). They provided colour images over a Gigabit Ethernet interface at a data rate of 1\,\unit{Gb\,s^{-1}} and supported a maximum frame rate of 43\,\unit{Hz}. The exposure time and the frame-rate were adjusted before the start of the missions in order to achieve the optimal horizontal/vertical overlap between the stereo pair, and to minimise the motion blur while simultaneously maintaining a wide range of intensities across the entire image.

The camera intrinsics were calculated by placing a ChArUco board underwater. A ChArUco board is a combination of a chessboard with ArUco markers \citep{ArucoGarrido} instead of the white squares. The ArUco markers are easily identifiable and their positions inside the chessboard are known beforehand, thus it is not necessary to frame the entire board inside the image bounds in order to perform the calibration. The water introduces additional image distortions to the already present lens distortion, however those distortions are mostly radial in nature, therefore the intrinsic camera calibration will be able to model these distortion with higher radial distortion coefficients. The camera calibration parameters $(f_x, f_y, x_0, y_0)$ and the radial $(k_1, k_2, k_3)$ and tangential lens distortion $(p_1, p_2)$ parameters are listed in Table~\ref{tab:intrinsics}.

\begin{table}[!h]
    \small\sf\centering
    \caption{Camera intrinsic parameters and lens distortion coefficients.}
    \begin{tabular}{ll}
        \hline
        \textbf{Parameter} & \textbf{Value} \\
        \hline
        \textbf{Camera Intrinsics} \\
        $f_x$ & 1801.899\,\unit{px} \\
        $f_y$ & 1801.899\,\unit{px} \\
        $x_0$ & 938.805\,\unit{px} \\
        $y_0$ & 735.677\,\unit{px} \\
        \textbf{Lens Distortion} \\
        $k_1$ & -0.19027 \\
        $k_2$ & 0.10212 \\
        $p_1$ & 0.00139 \\
        $p_2$ & -0.00343 \\
        $k_3$ & 0.00342 \\
        \hline
    \end{tabular}
    \label{tab:intrinsics}
\end{table}

The dimensions of the camera's sensor are $8.71\times6.588$\,\unit{mm}. The focal length of the camera (assuming a pinhole camera model) was calculated as 8.11\,\unit{mm}. The horizontal and vertical angles of view based on this focal length were calculated, respectively, as 68.26\,\unit{\degree} and 49.29\,\unit{\degree}.

During the survey, the orientation of the stereo pair was modified to better suit the image overlap and the visible information inside the field of view of the cameras. The extrinsic calibration was performed offline after the missions were completed. Table~\ref{tab:image_overlap} shows the percentage overlap and angle between the stereo pair for each day of the survey. 

\begin{table}[!h]
    \small\sf\centering
    \caption{Percentage overlap and angle between stereo pair for each day of the surveys.}
    \begin{tabular}{lll}
        \hline
        \textbf{Date} & \textbf{Overlap} & \textbf{Angle} \\
        \hline
        $22/04/2024$ & 26.34\,\% & 26.62\,\unit{\degree} \\
        $25/04/2024$ & 22.73\,\% & 28.62\,\unit{\degree} \\
        $26/04/2024$ & 25.2\,\% & 27.26\,\unit{\degree}\\
        \hline
    \end{tabular}
    \label{tab:image_overlap}
\end{table}

Since the extrinsic calibration was not performed using a calibration pattern, and the stereo pair is only offset by a known baseline and angle, it is advisable to optimise them using the provided extrinsics as initial estimates when employing the stereo pairs for SLAM or localisation tasks.

\subsection{Data Processing}
\label{sec:data_processing_o}

The navigation and optical data was recorded on a ROS bag \citep{ROSBag}, a file format in ROS \citep{RobotOperatingSystem} for storing ROS messages. Bags store serialised message data so it can replayed back in the exact way there were recorded. The navigation data included the altitude data and the pose of the robot from the Phins Compact C3 INS, while the optical data consisted of the images from the two Blackfly S cameras. The optical data was recorded in a compressed format to reduce the memory size of each bag recorded onboard the Girona1000's computer. The images were saved with a 90\,\% JPEG compression percentage in order to maintain a very high-quality image while gaining a significant reduction on the original 100\,\% file. No further post-processing was applied on the optical data. Upon decompressing the images, the average image size was 4.5\,\unit{MB}.

The Phins Compact C3 INS inside the Girona1000 improves the navigation drift by employing all the navigation sensors in its UKF. The Girona1000 navigation, together with the camera extrinsics (Tables \ref{tab:girona_sensor_distance_offsets} and \ref{tab:girona_sensor_angle_offsets}), was employed to calculate the exact position of each of the two cameras at every time instance. In order to assign the correct navigational information to the images, the image timestamps were matched to the navigation timestamps, and thus the corresponding navigational data was assigned to the temporally-closest image. 

However, the camera images would not have been usable at the altitude at which the SSS data was captured. The effects of the water column on light propagation (refraction, light absorption, light scattering, and colour shift) cause severe image quality degradation and distortion \citep{ShuImageEnhancement}. As a consequence, we surveyed the same area that had previously been covered during the acoustic surveys, but at a lower altitude. Additionally, to prevent the inclusion of degraded images in the dataset, we filtered out the images that were captured above 5\,\unit{m} of altitude from the seabed during the post-processing. This resulted in approximately 178000 images acquired from both cameras.


\section{Processing and Compilation of Training Datasets}
\label{sec:dataset}

\subsection{SSS Datasets}
\label{sec:dataset_a}

Mapping the benthic habitat in mission-time using AUVs necessitates direct processing of the raw sensor readings as they are acquired from the SSS. The raw 12-bit waterfalls were first log-scaled and normalized to the [0,1] range to ensure compatibility with ML pipelines, particularly for training DNNs. We used the natural logarithm (\texttt{log1p}) instead of the conventional base-10 logarithm (\texttt{log10}) to preserve numerical precision in low-intensity regions. This transformation improves the dynamic range compression while avoiding undefined values for zero inputs, which are common in sonar returns. The normalized log-transformed waterfalls were stored as NumPy arrays \citep{numpy}. Subsequently, slant range correction was applied to the waterfalls, and the available navigation data was used to geocode each bin. This allowed us to fetch the corresponding annotations from the ArcGIS interpretations and automatically generate ground truth for the waterfalls.

\begin{figure}[!b]
    \centering
    \includegraphics[width=0.47\textwidth]{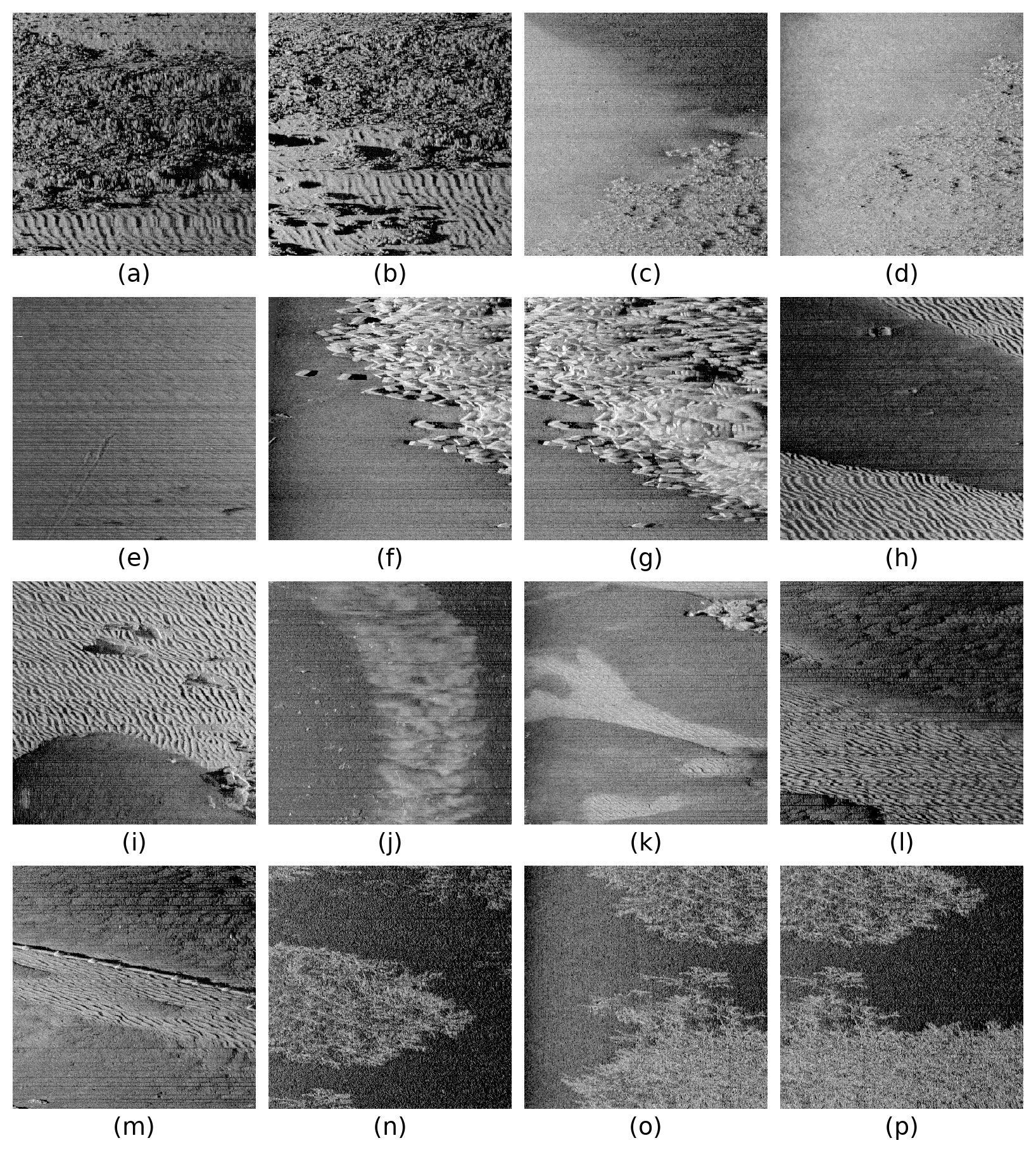}
    \caption{Samples of SSS tiles generated from sector N07 waterfalls.}
    \label{fig:sss_tiles}
\end{figure}

However, the original annotations on the mosaics were done on a much coarser resolution than required for accurate pixel-wise supervision on the waterfalls, which caused the generated segmentation masks to suffer from issues such as unclear inter-class boundaries and absence of annotations for small-scale objects. Additionally, unknown parameters for the delineation process used for mosaicing further introduced spatial discrepancies. Moreover, due to the sheer volume of the data, it was not feasible to produce well curated annotations for the entire dataset. Therefore, a subset of approximately 750 waterfalls, each composed of 1152 pings, was selected across different sectors that provided a suitable distribution of benthic classes. From this subset, 600 waterfalls originating from sectors \textit{N01}, \textit{N02}, \textit{N04}, \textit{N07}, \textit{N08}, \textit{S01}, \textit{S02} and \textit{S03} comprised the training set, while 150 waterfalls from sectors \textit{N03}, \textit{N05}, \textit{N06},  \textit{N09}, \textit{N10}, \textit{N13}, \textit{N15} and \textit{N16} were set aside for the test set. A Python-based annotation tool\footnote{\url{https://github.com/CIRS-Girona/MILF}} was developed to support manual refinement of the transferred segmentation masks. Appendix~\ref{sec:appendixA} provides a detailed breakdown of class-wise area coverage, sector-wise distributions, variations in survey altitude and depth, along with other data statistics.

The SSS records intensities of the acoustic backscatter in a series of parallel swaths of data, stacking successive swaths on top of each other as the AUV continues scanning lines along its direction of motion. The so formed acoustic waterfalls can, however, get too large to be processed in a single pass for each transect of the survey. Therefore, these waterfalls are usually partitioned in batches of certain lines in order to generate smaller tiles, each with a certain degree of overlap along-track and across-track so as to maintain continuity of information. Each tile can then be treated as an independent input to the classification algorithm and the results can be merged back into a coherent waterfall \citep{rtseg, dcnet, burguera, convit}.

\begin{figure}[!ht]
    \centering
    \includegraphics[width=0.47\textwidth]{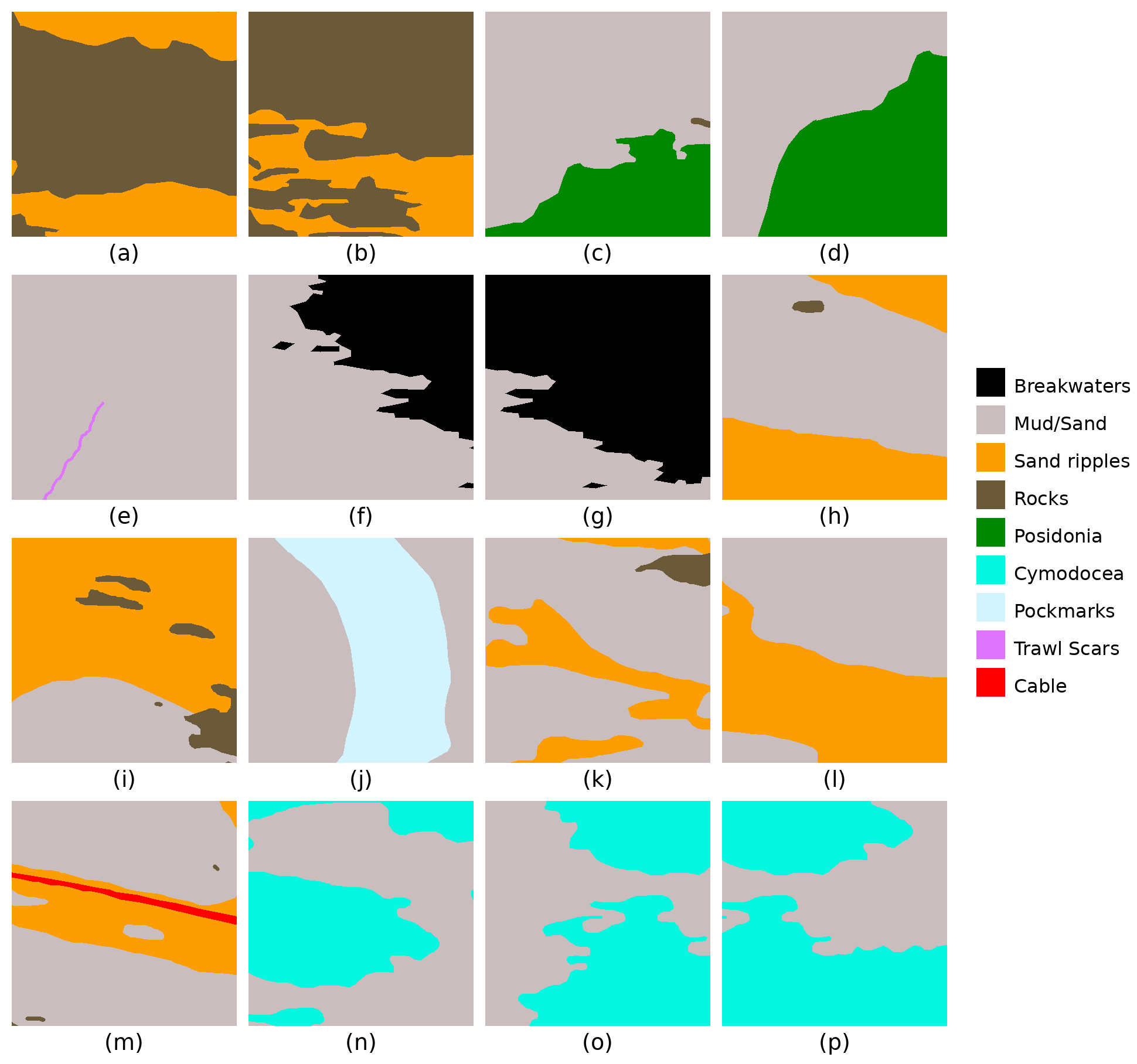}
    \caption{Segmentation masks corresponding to the SSS tiles from sector N07 shown in Fig.~\ref{fig:sss_tiles}.}
    \label{fig:segmentation_masks}
\end{figure}

In order to generate the datasets presented in this work, the waterfalls from all surveyed sectors were partitioned in batches of 384 pings to generate tiles of size $384\times384$, with a 192\,\unit{px} overlap along-track and across-track. This yielded over 950000 tiles. Figure~\ref{fig:sss_tiles} depicts some examples of these tiles. This large volume of unlabelled data can be leveraged for self-supervised pre-training of classification models, thereby serving as a strong proxy to learn semantically meaningful embeddings of the input. For supervised fine-tuning, we similarly partitioned the corrected segmentation masks obtained from the transferred annotations. This resulted in a approximately 36000 tiles with corresponding pixel-wise annotated ground truth, of which about 30000 were designated to the training set, while about 6000 were reserved for the test set. Figure~\ref{fig:segmentation_masks} illustrates some examples of the data.

\subsection{Multi-modal Dataset}
\label{sec:dataset_m}

As previously discussed, integrated surveys of the seafloor can substantially benefit from the fusion of information from multiple sensors. However, training ML models on such data poses inherent challenges with respect to the alignment between modalities. In our case, the acoustic surveys were more extensive, covering a significantly large area, whereas the optical surveys were more selective, targeting regions containing certain classes of interest. Consequently, there was no direct spatial alignment between the optical and acoustic data, with many SSS transects lacking corresponding optical imagery. Moreover, even where both modalities were available, the SSS tiles spanned much larger spatial footprints compared to the field of view of individual optical images.

To address this challenge, we adopted a geo-referencing strategy that associates multiple optical images with their corresponding SSS tiles based on spatial proximity. The resulting opti-acoustic groupings do not require strict pixel-level alignment and can be used to facilitate self-supervised learning of cross-modal feature representations via pretext tasks, such as visual similarity matching, wherein the model learns to associate acoustically and optically similar seafloor characteristics without relying on explicit labels. This strategy not only introduces meaningful optical context into SSS classification but also circumvents the need for manual annotation of optical images. Incorporating such visual context can provide an effective mechanism to reduce class ambiguities in acoustic imagery, particularly in scenarios where distinct benthic classes exhibit similar backscatter characteristics but visually distinct textures or colours.


The opti-acoustic registration process involves three main steps:

\begin{enumerate}
    \item Calculating the geographical coordinates of SSS tiles
    \item Calculating the geographical coordinates of optical images
    \item Computing the intersection of the optical and SSS footprints on the seabed
\end{enumerate}

The geographical coordinates $\left( x_j^i, y_j^i  \right) \in \mathbb{R}^2$ of bin $j$ inside SSS ping $i$ were computed using the following set of equations (assuming a flat seabed):

\begin{align}
    x_j^i &= x_i + \Delta x_j \cos{\psi_i} + \Delta y_i \sin\psi_i \label{eq:sonar_footprint_x} \\
    y_j^i &= y_i - \Delta x_j \sin{\psi_i} + \Delta y_i \cos\psi_i  \label{eq:sonar_footprint_y} \\
    \Delta x_j &= \sqrt{\left( \frac{j \cdot r}{n} \right)^2 - \left( \frac{h_i}{\cos{\theta_i}} \right)^2} \\
    \Delta y_i &= h_i \tan{\theta_i}
\end{align}

\begin{figure*}[!ht]
    \centering
    \includegraphics[width=0.96\textwidth]{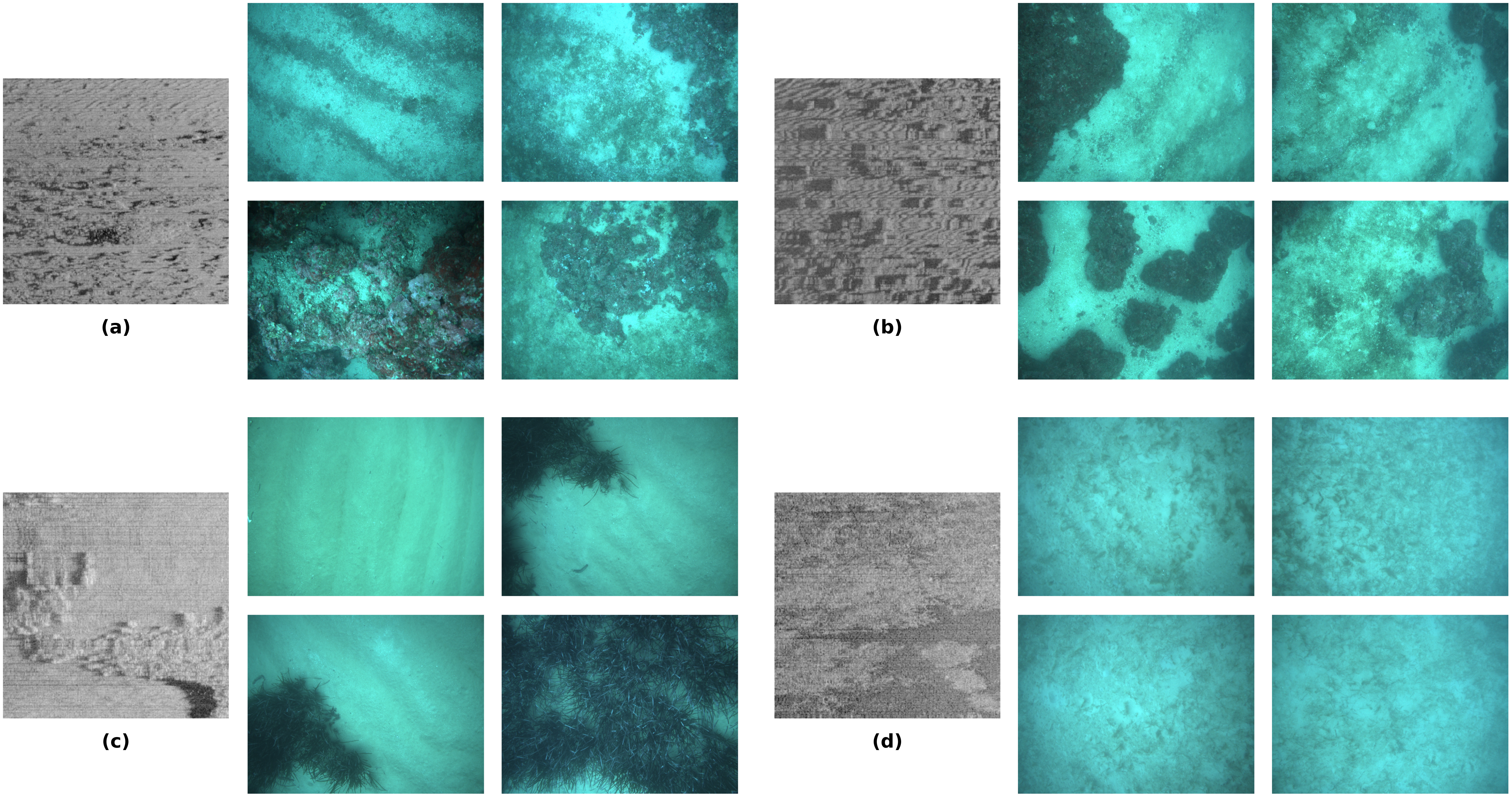}
    \caption{Examples of SSS tiles from sectors N07 and N08, alongside a subset of their corresponding matched optical images.}
    \label{fig:opto_acoustic_data}
\end{figure*}

\noindent where $x_i \in \mathbb{R}$ and $y_i \in \mathbb{R}$ denote the easting and northing coordinates at ping $i$, $\psi_i$ and $\theta_i$ are the yaw and pitch rotations at ping $i$, $h \in \mathbb{R}$ is the altitude at ping $i$, $n$ is the number of bins inside ping $i$, and $r \in \mathbb{R}$ is the slant resolution. The pitch $\theta_i$ angle helps determine the width and the offset geographical coordinates of the SSS ping, since a pitching movement will cause the SSS pulses to travel further away.

The top-left, top-right, bottom-right and bottom-left bins of an SSS tile (composed of multiple pings) can be used to calculate the bounds of the spatial footprint of the SSS on the seabed via Eq.~(\ref{eq:sonar_footprint_x}) and Eq.~(\ref{eq:sonar_footprint_y}).

Similarly, the geographical coordinates $\mathbf{p}^W \in \mathbb{R}^3$ of the projection of a pixel $\mathbf{p}^I \in \mathbb{R}^2$ of an image on the seabed are determined by:

\begin{equation}
    \mathbf{p}^W = {}^w\mathbf{T}_{c}  \mathbf{K}^{-1} \cdot h \cdot \mathbf{p}^I
\label{eq:image_footprint}
\end{equation}

\noindent where $\mathbf{K}$ denotes the camera intrinsic matrix, ${}^w\mathbf{T}_c$ represents the transformation matrix from the world to camera frame. Both the intrinsic and extrinsic matrices are required for this operation, and are stored inside a .json file. This camera calibration metadata file contains dictionary objects where each entry corresponds to a distinct ROS Bag recording. Each dictionary entry stores the ROS topic name and the intrinsic and extrinsic parameters of the left and right cameras used during the optical surveys. 

The top-left, top-right, bottom-left and bottom-right corners of an image can be used to calculate the bounds of the optical image footprint on the seabed via Eq.~(\ref{eq:image_footprint}), assuming a flat seabed.

A matching procedure was then performed on the seabed footprints of both acoustic and optical data: if more than 40\,\% of an image lies within the bounds of an SSS patch, the pair is considered a \textit{match}. The process was repeated for all the SSS tiles and optical images until the entire dataset was constructed. SSS tiles with no optical matches were ignored.

Once the matching procedure was completed, three metadata files were generated:
\begin{itemize}
    \item SSS tile metadata (.csv file) 
    \item Optical image metadata (.csv file)
    \item Opti-acoustic registration metadata (.json file)
\end{itemize}

The opti-acoustic registration metadata consists of dictionary objects, defined as unordered collections of values with unique keys. Each key represents the index of an SSS tile, and the corresponding value is a list of indices identifying the optical images associated with that tile based on the matching criteria. A summary of the SSS and camera metadata files is provided in Appendix~\ref{sec:appendixB}.

For the final dataset, we reduced the size of SSS tiles to $256 \times 256$ pixels to better balance the spatial scale between acoustic and optical modalities, minimizing the discrepancy in seabed footprint covered by each data sample. This resulted in a total of 2080 SSS tiles and 35947 optical images, yielding an average of approximately 18 optical images associated with each SSS tile. Each SSS tile corresponds to a seabed footprint of approximately 25\,\unit{m^2}, whereas individual optical images cover between 11\,\unit{m^2} and 18\,\unit{m^2}, depending on vehicle altitude and camera configuration. Figure~\ref{fig:opto_acoustic_data} depicts an SSS tile and a few of the associated optical images.



\section{Repository Structure}
\label{sec:repository}

The data is provided in both raw and processed formats so as not to constrain users to a specific application. It has also been synchronised and timestamped to make it easily accessible. It is organised as a collection of individual datasets. The collection currently comprises four datasets: 

\begin{itemize}
    \item \textbf{BenthiCat - Raw}, which contains all the unprocessed files, including ROS Bag recordings (together with the associated calibration metadata) from the optical imaging surveys, XTF files from the SSS data, and \texttt{.xyz} files from the MBES data. The \texttt{SSS} and \texttt{MBES} folders are further subdivided into sector-specific subdirectories, each containing the corresponding raw data files for that sector. Additionally, the \texttt{GIS} folder contains the georeferenced mosaics and ArcGIS interpretations. Figure~\ref{fig:repo_structure_raw} presents an overview of the repository structure.

    \begin{figure}[!h]
        \centering
        \includegraphics[width=0.47\textwidth]{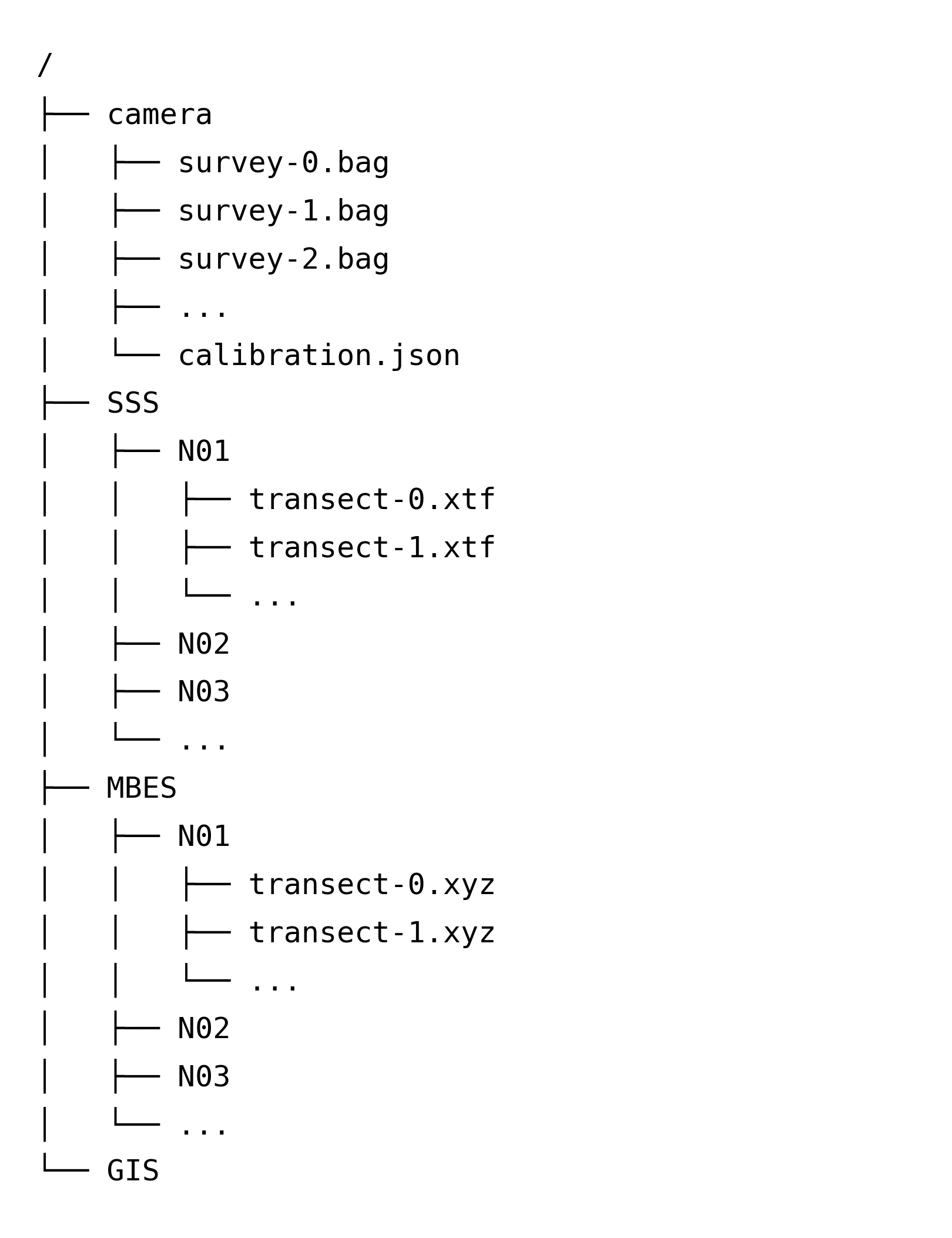}
        \caption{Structure of the raw dataset.}
        \label{fig:repo_structure_raw}
    \end{figure}
    
    \item \textbf{BenthiCat - SSS Pre-training} that contains unlabelled SSS tiles for unsupervised representation learning and self-supervised pre-training tasks.
    
    \item \textbf{BenthiCat - SSS Seafloor Segmentation} containing the manually annotated SSS waterfalls for supervised fine-tuning. The dataset is divided into \texttt{train} and \texttt{test} subsets, with the \texttt{test} set consisting of independent transects that do not spatially overlap with those in the \texttt{train} set.
    
    The dataset contains waterfalls instead of pre-cut tiles to avoid constraining users to a fixed tile size; however, a Python script to generate overlapping tiles from the waterfalls is available in the GitHub repository.
    
    \item \textbf{BenthiCat - SSS-CAM}, which contains the spatially matched optical and SSS samples. It includes two folders: \texttt{sonar} and \texttt{camera}, along with the three metadata files: \texttt{sonar.csv}, \texttt{camera.csv} and \texttt{correspondences.json}. The \texttt{sonar} and \texttt{camera} folders store all the matched SSS tiles and optical images, respectively. An overview of the repository structure is presented in Fig.~\ref{fig:repo_structure_oa}.

    \begin{figure}[!h]
        \centering
        \includegraphics[width=0.47\textwidth]{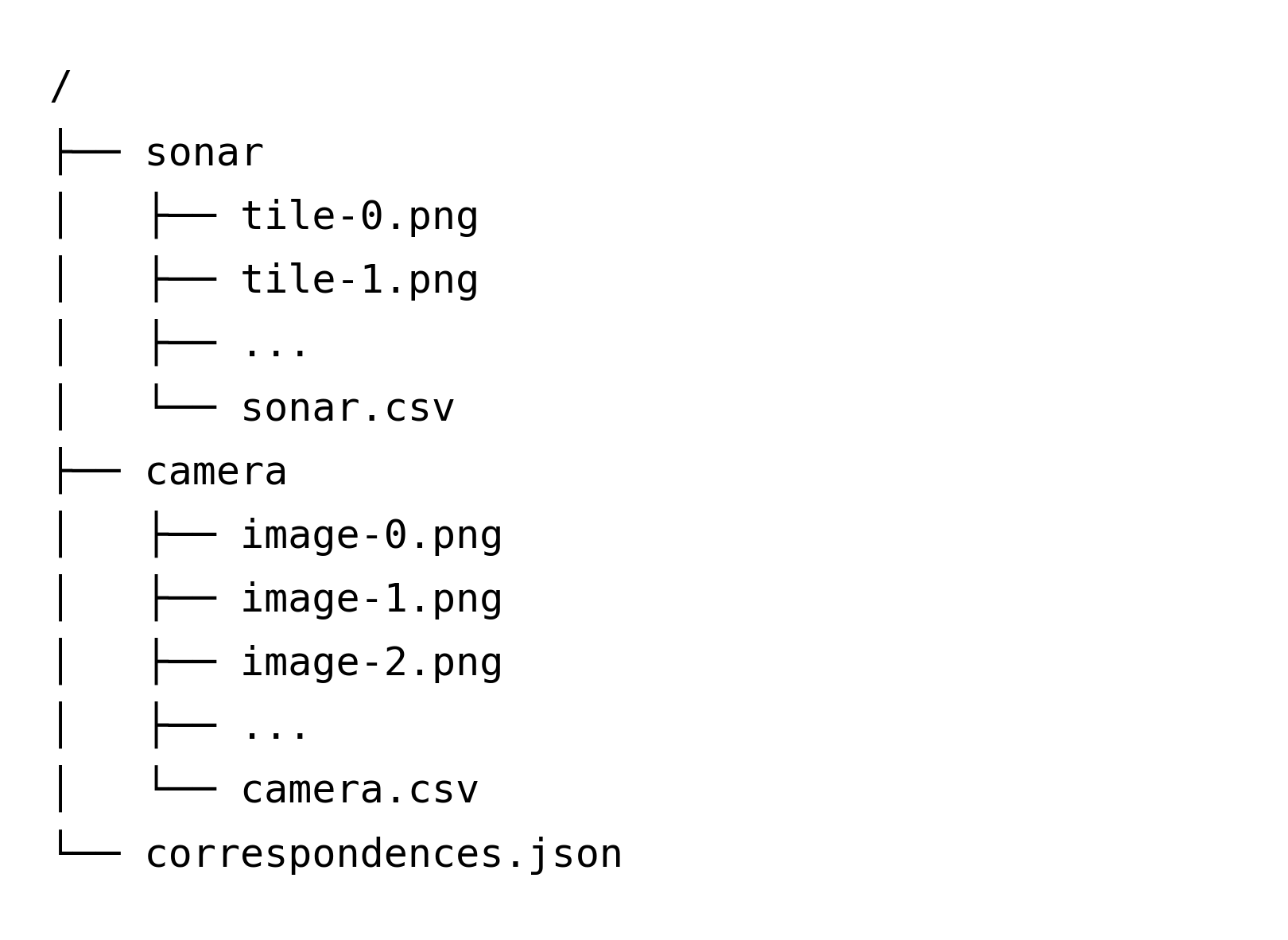}
        \caption{Structure of the opti-acoustic dataset.}
        \label{fig:repo_structure_oa}
    \end{figure}

\end{itemize}


\conclusions  
\label{sec:conclusion}

This work presents a comprehensive multi-modal dataset combining SSS and optical imagery for benthic habitat mapping using AUVs. The dataset includes both SSS-only and opti-acoustic subsets, with detailed metadata, curated annotations, and standardized training and testing splits designed to facilitate benchmarking ML algorithms. In addition to the curated datasets, we provide open access to the raw data, including \textit{rosbag} recordings, XTF files, and MBES outputs, enabling researchers to apply alternative processing strategies or generate task-specific datasets. Furthermore, all data preprocessing and annotation scripts are made openly available for reproducible research and to foster rapid extensibility.



\codedataavailability{The entire data collection is available on Harvard Dataverse under \url{https://dataverse.harvard.edu/dataverse/benthicat}. Exemplar code and detailed instructions on how to process the data are available at \url{https://github.com/CIRS-Girona/opti-acoustic-preprocessing-tools}.}


\appendix

\section{Supplementary Dataset Statistics}
\label{sec:appendixA}

This appendix provides supplementary statistics and analyses of the datasets presented in this study. The tables included herein provide detailed information on class distributions, spatial coverage, and additional relevant metrics not reported in the main text. These materials are intended to support research and further exploration of the dataset.

Table~\ref{tab:classes} lists the classes selected for annotation within the SSS dataset, along with their assigned IDs and the RGB triplets used for visualizing the annotations.

\begin{table}[!h]
    \small\sf\centering
    \caption{Summary of annotated classes in the SSS dataset.}
    \begin{tabular}{clcc}
        \hline
        \textbf{ID} & \textbf{Class} & \textbf{RGB} \\
        \hline
        0 & undefined & (255,255,255) \\
        1 & breakwaters & (0,0,0) \\
        2 & artificial\_reefs & (57,48,138) \\
        3 & rocks & (107,90,58) \\
        4 & corals & (127,14,138) \\
        5 & sand\_ripples & (252,140,4) \\
        6 & fine\_sediment & (235,186,101) \\
        7 & cymodocea & (5,247,215) \\
        8 & posidonia & (0,136,0) \\
        9 & cables & (255,0,0) \\
        10 & trawl\_scars & (223,116,255) \\
        11 & misc\_objects & (140,129,247) \\
        \hline
    \end{tabular}
    \label{tab:classes}
\end{table}

\begin{sidewaysfigure}[!h]
    \centering
    \includegraphics[width=0.96\textwidth]{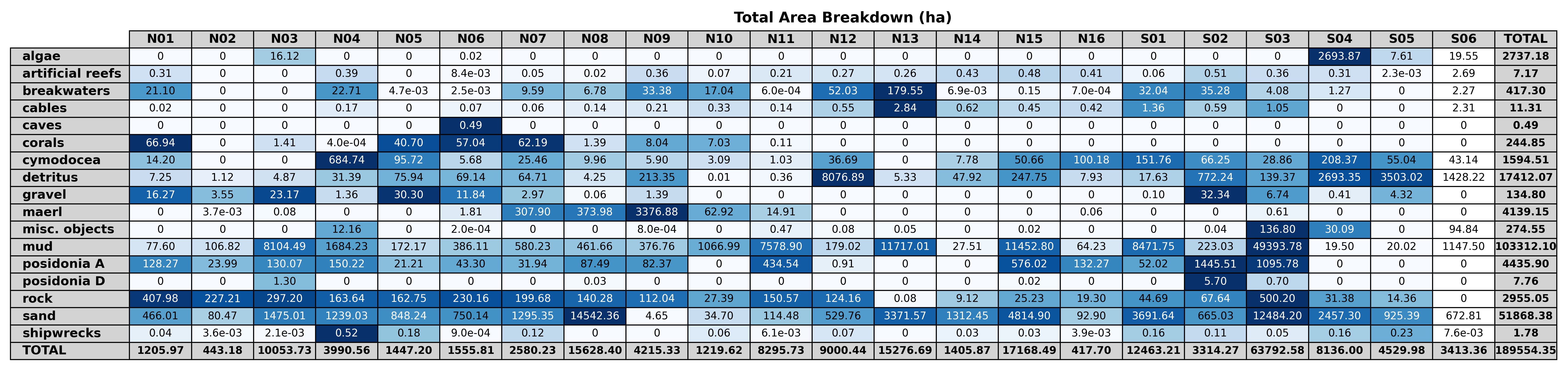}
    \caption{Absolute area (in hectares) of each class within each sector.}
    \label{fig:area_breakdown}
\end{sidewaysfigure}

\begin{sidewaysfigure}[!h]
    \centering
    \includegraphics[width=0.96\textwidth]{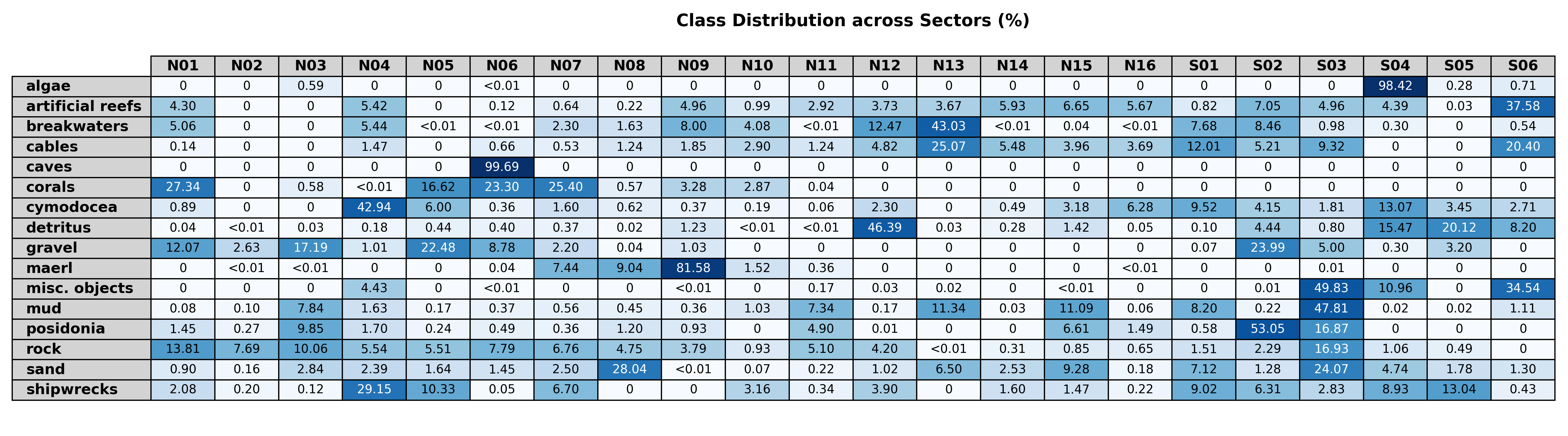}
    \caption{Distribution of each class across all sectors. Values are expressed as the percentage of the total area of each class contributed by every sector.}
    \label{fig:sector_distribution}
\end{sidewaysfigure}

\begin{sidewaysfigure}[!h]
    \centering
    \includegraphics[width=0.96\textwidth]{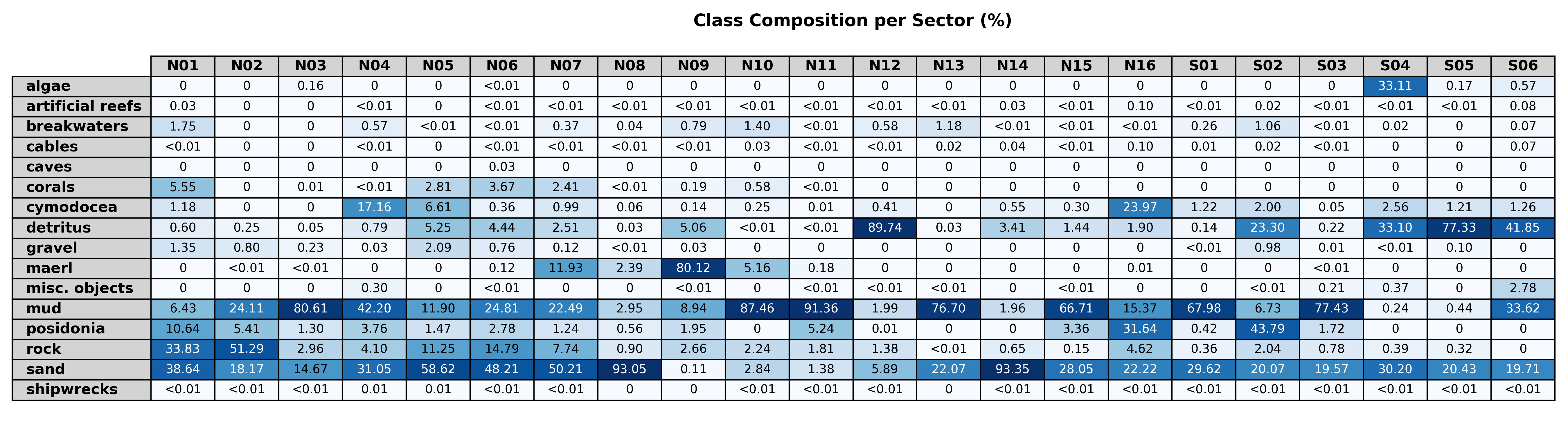}
    \caption{Relative composition of classes within each sector. Values are expressed as the percentage of the total annotated area per sector.}
    \label{fig:sector_composition}
\end{sidewaysfigure}

Figures~\ref{fig:area_breakdown}\textendash\ref{fig:sector_composition} summarize the distribution of benthic classes across all surveyed sectors. Figure~\ref{fig:sector_distribution} shows the distribution of each class across all sectors, expressed as the percentage of the total area of each class contributed by every sector, whereas Fig.~\ref{fig:sector_composition} shows the relative contribution of each class within individual sectors, expressed as a percentage of the total annotated area per sector. Figure~\ref{fig:area_breakdown}, on the other hand, presents the absolute area (in hectares) of each class within every sector.

\begin{figure*}[!h]
    \centering
    \includegraphics[height=0.96\textheight,keepaspectratio]{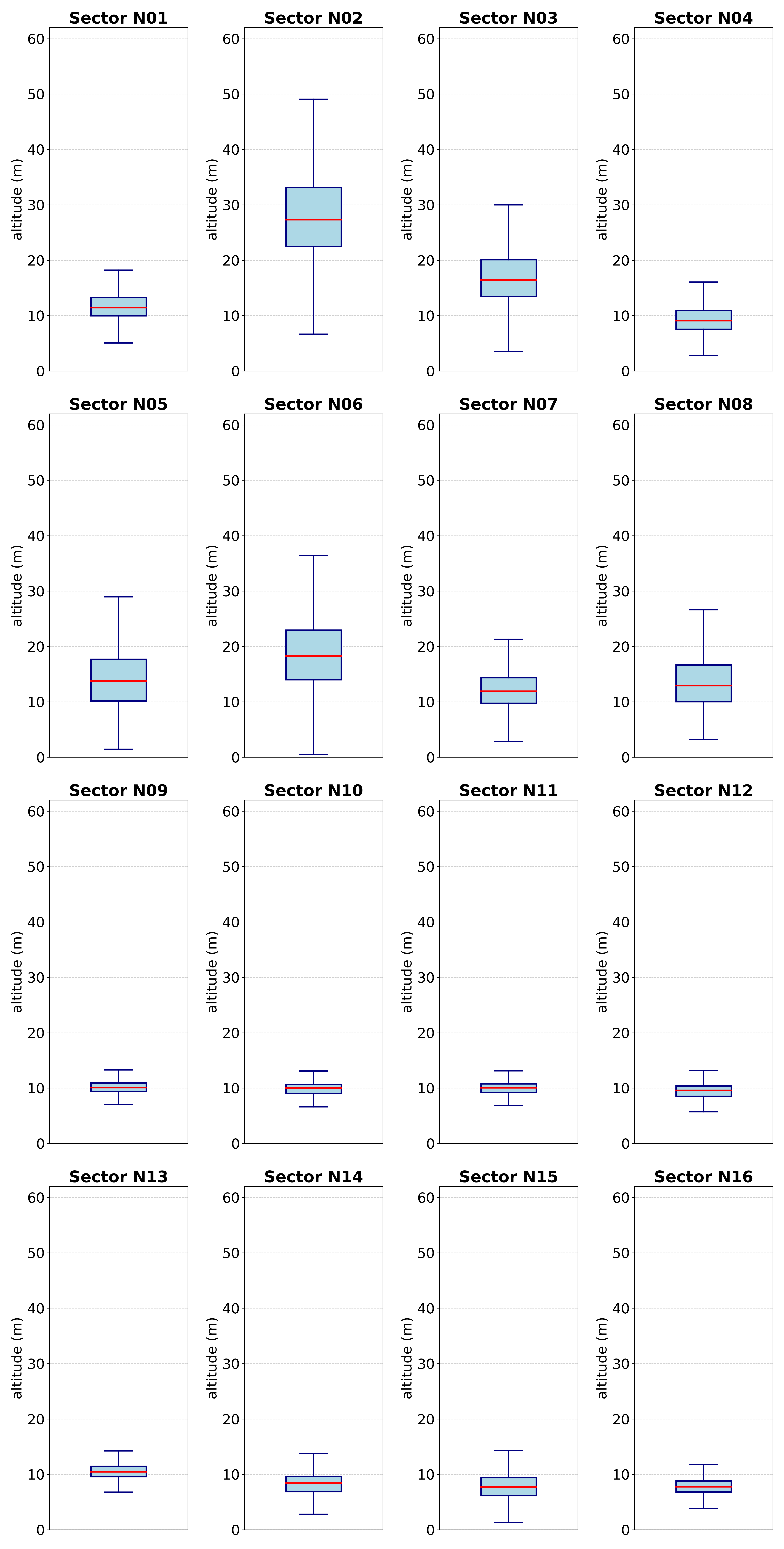}
    \caption{Box plots of survey altitudes for the SSS campaigns in the northern sectors.}
    \label{fig:altitudes_N}
\end{figure*}

\begin{figure*}[!h]
    \centering
    \includegraphics[height=0.42\textheight,keepaspectratio]{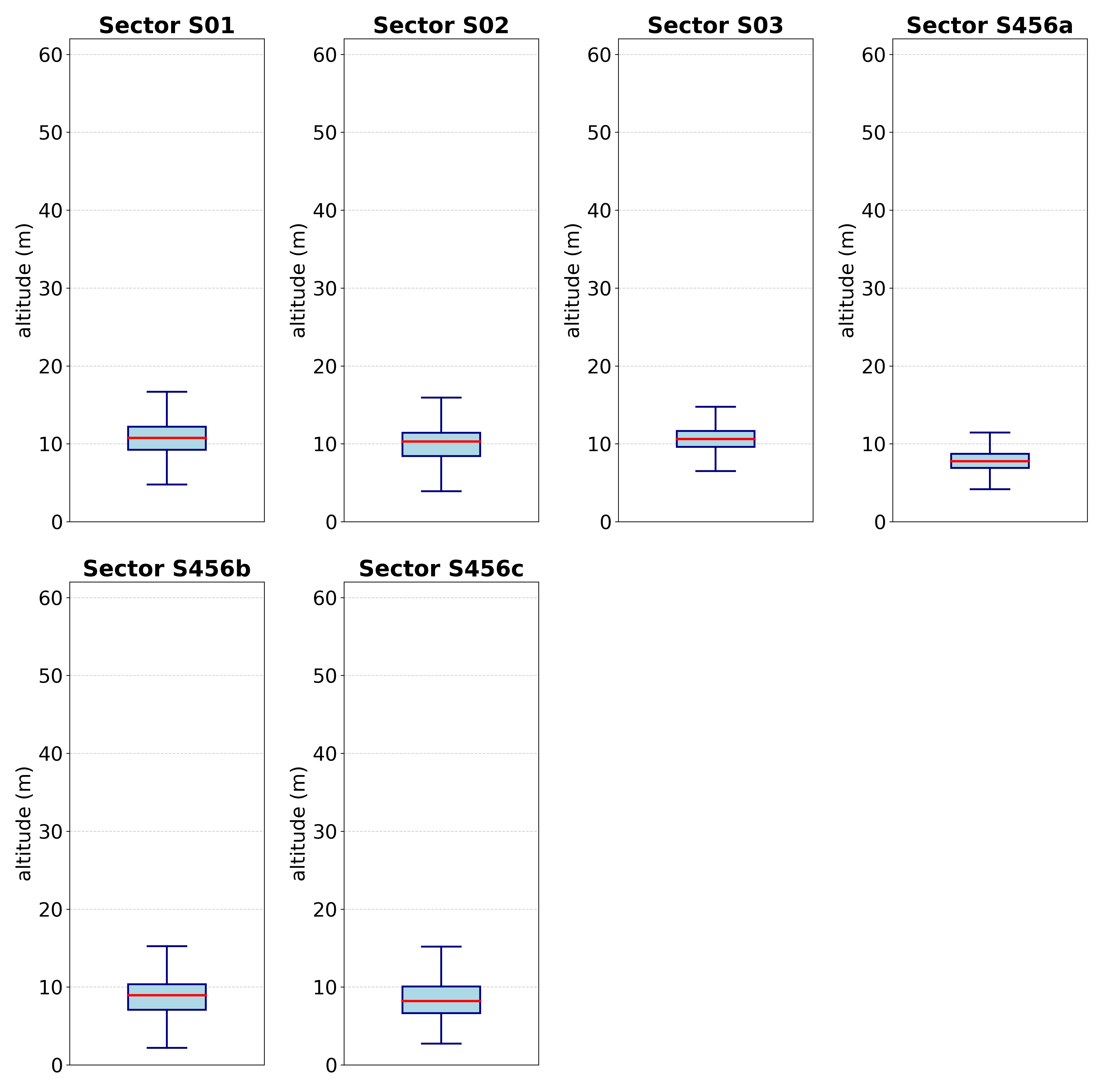}
    \caption{Box plots of survey altitudes for the SSS campaigns in the southern sectors.}
    \label{fig:altitudes_S}
\end{figure*}

\begin{figure*}[!h]
    \centering
    \includegraphics[height=0.42\textheight,keepaspectratio]{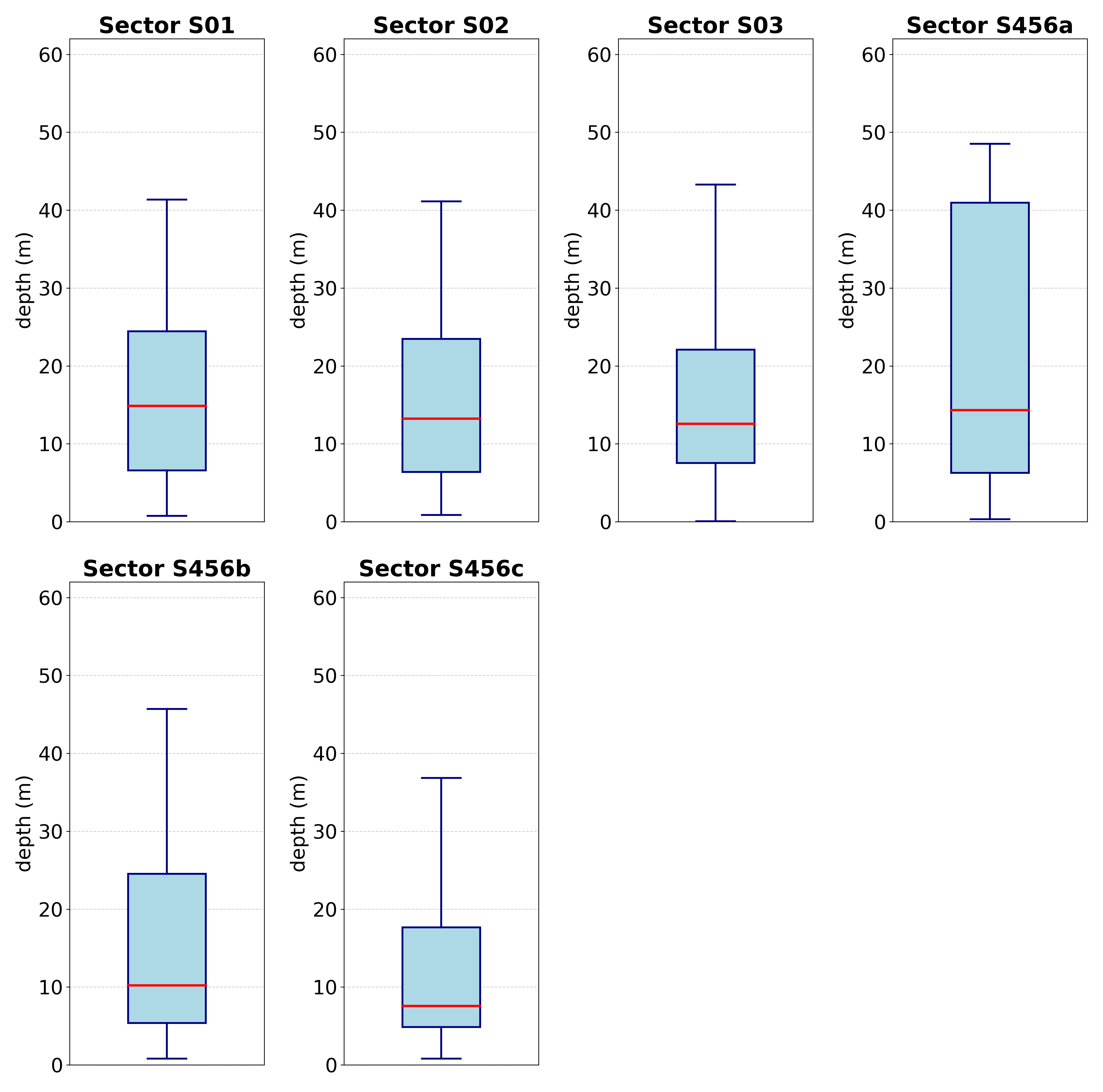}
    \caption{Box plots of survey depths for the SSS campaigns in the southern sectors.}
    \label{fig:depths_S}
\end{figure*}

\begin{figure*}[!h]
    \centering
    \includegraphics[height=0.96\textheight,keepaspectratio]{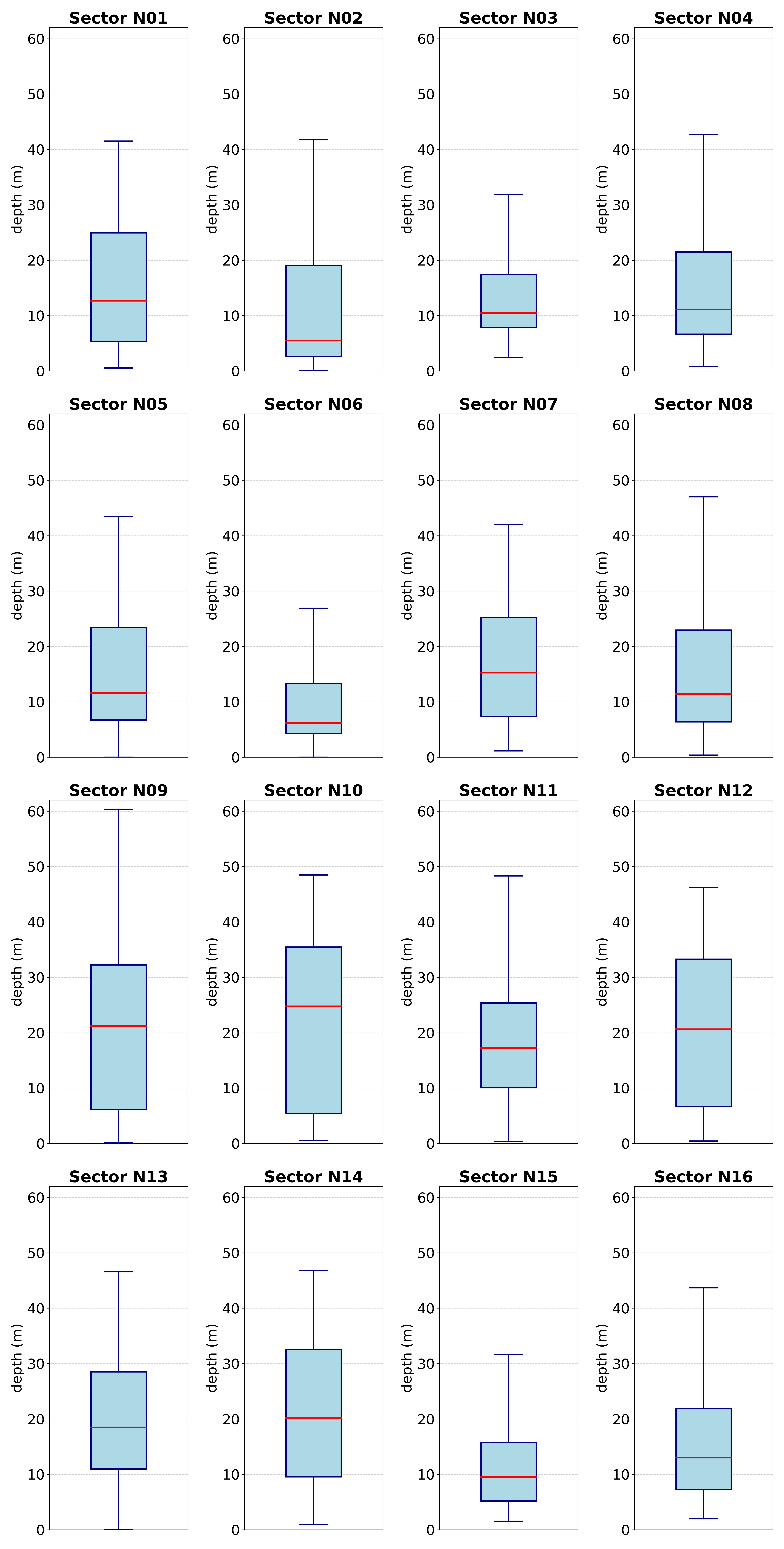}
    \caption{Box plots of survey depths for the SSS campaigns in the northern sectors.}
    \label{fig:depths_N}
\end{figure*}

Figures \ref{fig:altitudes_N} and \ref{fig:altitudes_S} illustrate the altitudes at which SSS surveys were conducted as box plots. The northern sectors are shown in Fig.~\ref{fig:altitudes_N} and the southern sectors in Fig.~\ref{fig:altitudes_S}. Similarly, figures \ref{fig:depths_S} and \ref{fig:depths_N} present the corresponding distribution of survey depths, with Fig.~\ref{fig:depths_S} for the southern sectors and Fig.~\ref{fig:depths_N} for the northern sectors.

\begin{figure*}[!h]
    \centering
    \includegraphics[width=0.96\textwidth]{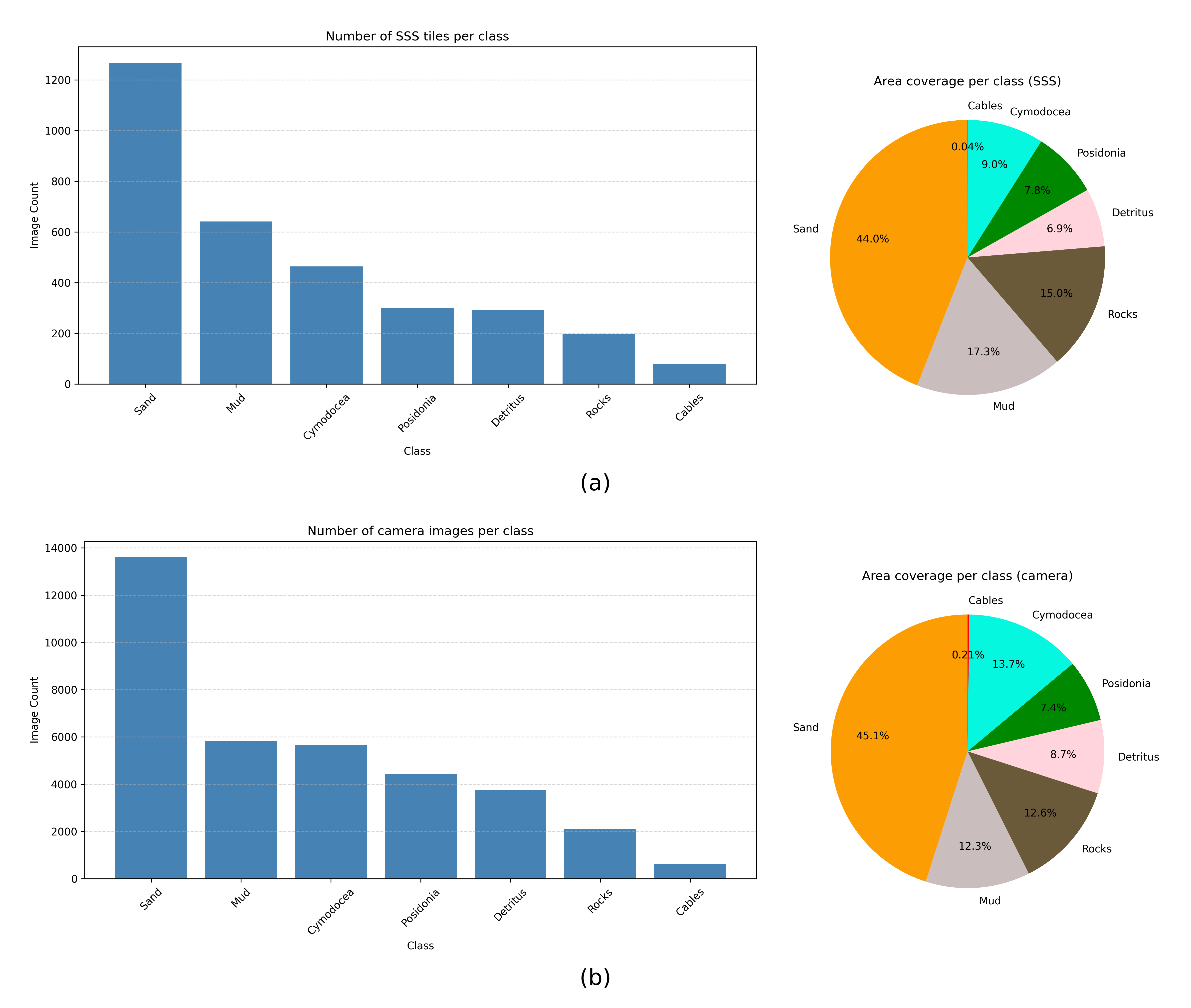}
    \caption{Class distribution within the multimodal dataset for (a) SSS tiles and (b) optical images.}
    \label{fig:class_distribution_oa}
\end{figure*}

Figure~\ref{fig:class_distribution_oa} summarizes the class distribution within the multimodal dataset in terms of the percentage of total area covered per class and the number of samples per class, for both optical images and SSS tiles.

\clearpage

\section{Metadata for Opti-acoustic Dataset}
\label{sec:appendixB}

This appendix provides a summary of metadata corresponding to the SSS tiles and camera images generated during the opti-acoustic matching procedure.

\begin{table}[!h]
    \small\sf\centering
    \caption{Metadata summary for SSS tiles.}
    \begin{tabular}{lll}
        \hline
        \textbf{Parameter} & \textbf{Unit} & \textbf{Data Type}\\
        \hline
        Timestamp & \unit{s} &\texttt{float} \\
        Patch file path & & \texttt{string} \\
        XTF file path & & \texttt{string} \\
        Start ping index & & \texttt{int} \\
        End ping index & & \texttt{int} \\
        Start bin index & & \texttt{int} \\
        End bin index & & \texttt{int} \\
        Easting &\unit{m} & \texttt{float} \\
        Northing &\unit{m} & \texttt{float} \\
        Roll & \unit{rad} &  \texttt{float}  \\
        Pitch & \unit{rad} &  \texttt{float} \\
        Yaw & \unit{rad} &  \texttt{float} \\
        Longitude & \unit{\degree} &  \texttt{float}  \\
        Latitude & \unit{\degree} &  \texttt{float}  \\
        EPSG code of UTM zone & & \texttt{int} \\
        Altitude & \unit{m} & \texttt{float} \\
        Top-left corner easting & \unit{m} & \texttt{float} \\
        Top-left corner northing & \unit{m} & \texttt{float} \\
        Top-right corner easting & \unit{m} & \texttt{float} \\
        Top-right corner northing & \unit{m} & \texttt{float} \\
        Bottom-right corner easting & \unit{m} & \texttt{float} \\
        Bottom-right corner northing & \unit{m} & \texttt{float} \\
        Bottom-left corner easting & \unit{m} & \texttt{float} \\
        Bottom-left corner northing & \unit{m}& \texttt{float}\\
        \hline
    \end{tabular}
    \label{tab:metadata_sss}
\end{table}

\begin{table}[!h]
    \small\sf\centering
    \caption{Metadata summary for camera images.}
    \begin{tabular}{lll}
        \hline
        \textbf{Parameter} & \textbf{Unit} & \textbf{Data Type}\\
        \hline
        Timestamp & \unit{s} & \texttt{float}\\
        Image file path & & \texttt{string} \\
        ROS Bag file path & & \texttt{string} \\
        ROS Topic name & & \texttt{string} \\
        ROS Message sequence index & & \texttt{int} \\
        Easting &\unit{m} & \texttt{float} \\
        Northing &\unit{m} & \texttt{float} \\
        Roll & \unit{rad} &  \texttt{float}  \\
        Pitch & \unit{rad} &  \texttt{float} \\
        Yaw & \unit{rad} &  \texttt{float} \\
        Longitude & \unit{\degree} &  \texttt{float}  \\
        Latitude & \unit{\degree} &  \texttt{float}  \\
        EPSG code of UTM zone & & \texttt{int} \\
        Altitude & \unit{m} & \texttt{float} \\
        Depth & \unit{m} & \texttt{float}\\
        Top-left corner easting & \unit{m} & \texttt{float} \\
        Top-left corner northing & \unit{m} & \texttt{float} \\
        Top-right corner easting & \unit{m} & \texttt{float} \\
        Top-right corner northing & \unit{m} & \texttt{float} \\
        Bottom-right corner easting & \unit{m} & \texttt{float} \\
        Bottom-right corner northing & \unit{m} & \texttt{float} \\
        Bottom-left corner easting & \unit{m} & \texttt{float} \\
        Bottom-left corner northing & \unit{m}& \texttt{float}\\
        \hline
    \end{tabular}
    \label{tab:metadata_cam}
\end{table}

\noappendix       





\authorcontribution{H. Rajani: Methodology, Software, Data Curation, Formal Analysis, Writing - original draft; V. Franchi: Software, Data Curation, Writing - original draft; B. Valles: Data Curation, Writing - review \& editing; R. Ramos: Conceptualization, Project Administration, Methodology, Writing - review \& editing; R. Garcia: Project Administration, Supervision, Validation, Writing - review \& editing; N. Gracias: Project Administration, Supervision, Validation, Writing - review \& editing.} 

\competinginterests{The authors declared no potential conflicts of interest with respect to the research, authorship, and/or publication of this article.}

\begin{acknowledgements}
The study was supported by the Spanish government through projects ASSiST (PID2023-149413OB-I00) and IURBI (CNS2023-144688). We would also like to acknowledge the Ministry of Climate Action, Food and Rural Agenda of the Catalan government for granting permission to publicly disclose specific segments of the georeferenced data.
\end{acknowledgements}

\bibliographystyle{copernicus}
\bibliography{bibliography.bib}

\end{document}